\documentclass{article}

\newif\ifarxiv
\arxivtrue

\ifarxiv
  \usepackage[preprint]{neurips_2026}
\else
  \usepackage{neurips_2026}
\fi

\usepackage[utf8]{inputenc} 
\usepackage[T1]{fontenc}    
\usepackage{hyperref}       
\usepackage{url}            
\usepackage{booktabs}       
\usepackage{array}          
\usepackage{multirow}       
\usepackage{amsfonts}       
\usepackage{nicefrac}       
\usepackage{microtype}      
\usepackage[table]{xcolor}  
\usepackage{graphicx}
\usepackage{amsmath}
\usepackage{fontawesome5}    
\usepackage[ruled,vlined]{algorithm2e}  

\definecolor{citecolor}{HTML}{0071bc}
\hypersetup{
  colorlinks=true,
  citecolor=citecolor
}

\newif\ifclean
\cleantrue

\newcommand{\SX}[1]{\ifclean{#1}\else {\textcolor{cyan}{#1}}\fi}
\definecolor{codexblue}{HTML}{1E5AA8}

\newcommand{\codex}[1]{\ifclean{#1}\else {\textcolor{codexblue}{#1}}\fi}

\makeatletter
\newcommand{\compactfloatspacing}{%
  \setlength{\textfloatsep}{8pt plus 2pt minus 2pt}%
  \setlength{\floatsep}{8pt plus 2pt minus 2pt}%
  \setlength{\intextsep}{8pt plus 2pt minus 2pt}%
  \setlength{\@neuripsabovecaptionskip}{5pt}%
  \setlength{\@neuripsbelowcaptionskip}{1pt}%
  \setlength{\abovecaptionskip}{5pt}%
  \setlength{\belowcaptionskip}{1pt}%
}
\newcommand{\defaultfloatspacing}{%
  \setlength{\textfloatsep}{20pt plus 2pt minus 4pt}%
  \setlength{\floatsep}{12pt plus 2pt minus 2pt}%
  \setlength{\intextsep}{12pt plus 2pt minus 2pt}%
  \setlength{\@neuripsabovecaptionskip}{7pt}%
  \setlength{\@neuripsbelowcaptionskip}{0pt}%
  \setlength{\abovecaptionskip}{7pt}%
  \setlength{\belowcaptionskip}{0pt}%
}
\makeatother
\compactfloatspacing

\RestyleAlgo{ruled}
\SetAlgoNoLine
\DontPrintSemicolon
\SetAlFnt{\small}
\SetAlCapFnt{\small\bfseries}
\SetAlCapNameFnt{\small\bfseries}
\SetAlgoCaptionSeparator{.}
\SetAlCapSkip{4pt}
\SetKwInput{KwIn}{Input}
\SetKwInput{KwOut}{Output}

\SetCommentSty{algcommentfont}

\definecolor{cellgreen}{HTML}{e5f5e0}
\definecolor{cellred}{HTML}{FFDDE1}
\definecolor{cellgray}{HTML}{E8E8E8}

\definecolor{axispersons}{HTML}{4A9D5F}
\definecolor{axisobjects}{HTML}{4A7BB7}
\definecolor{axisbehavior}{HTML}{E8804C}
\definecolor{axisego}{HTML}{7E5DAA}
\newcommand{\promptaxislabel}[2]{\textcolor{#1}{\textbf{#2}}}
\newcommand{\promptitemlabel}[1]{\textcolor{black!85}{#1}}
\newcommand{\promptsteplabel}[1]{\textcolor{black!75}{\textbf{#1}}}
\newenvironment{promptquote}{%
  \par\smallskip
  \begin{list}{}{%
    \setlength{\leftmargin}{1em}%
    \setlength{\rightmargin}{0.5em}%
    \setlength{\topsep}{0pt}%
    \setlength{\partopsep}{0pt}%
    \setlength{\parsep}{0pt}%
    \setlength{\itemsep}{0pt}%
  }%
  \item\small\raggedright
}{%
  \end{list}
  \smallskip
}

\newcolumntype{N}{>{\centering\arraybackslash}p{1.1cm}}
\newcolumntype{E}{>{\centering\arraybackslash}p{1.6cm}}

\newcommand{\GlobalTableArrayStretch}{1.2}
\newcommand{\bencheval}{%
  \small
  \setlength{\tabcolsep}{4pt}%
  \renewcommand{\arraystretch}{\GlobalTableArrayStretch}%
}

\newcommand{\benchheader}{%
  \toprule
  \multirow{2}{*}{Context regime}
    & \multicolumn{2}{c}{\textcolor{axispersons}{\faUsers~Persons}}
    & \multicolumn{2}{c}{\textcolor{axisobjects}{\faWineBottle~Objects}}
    & \multicolumn{2}{c}{\textcolor{axisbehavior}{\faUtensils~Behavior}}
    & \textcolor{axisego}{\faUser~EgoWearer} \\
  \cmidrule(lr){2-3}\cmidrule(lr){4-5}\cmidrule(lr){6-7}\cmidrule(lr){8-8}
    & ID & Rel & ID & Det & Err & QA & ID \\
  \midrule
}

\newcommand{\benchmidrule}{\specialrule{\lightrulewidth}{0pt}{0pt}}
\newcommand{\benchbottomrule}{\specialrule{\heavyrulewidth}{0pt}{0pt}}

\title{Personal Visual Context Learning \\in Large Multimodal Models}

\ifarxiv
  \author{%
    Zihui Xue
    \quad Ami Baid
    \quad Sangho Kim
    \quad Mi Luo
    \quad Kristen Grauman \\
    The University of Texas at Austin 
  }
\else
  \author{%
    David S.~Hippocampus\thanks{Use footnote for providing further information
      about author (webpage, alternative address)---\emph{not} for acknowledging
      funding agencies.} \\
    Department of Computer Science\\
    Cranberry-Lemon University\\
    Pittsburgh, PA 15213 \\
    \texttt{hippo@cs.cranberry-lemon.edu} \\
  }
\fi

\begin{document}

\maketitle

\begin{abstract}

As wearable devices like smart glasses integrate Large Multimodal Models (LMMs) into the continuous first-person visual streams of individual users, the evolution of these models into true personal assistants hinges on visual personalization: the ability to reason over visual information unique to the wearer.
We formalize this capability as Personal Visual Context Learning (Personal VCL), the prompt-time capability of using user-specific visual context to resolve personalized queries. To systematically evaluate this, we present Personal-VCL-Bench, a comprehensive benchmark capturing the personal visual world across persons, objects, and behaviors. Our analysis of frontier LMMs identifies a profound context utilization gap, revealing that the mechanisms for leveraging visual evidence, as well as aggregating multiple visual observations, remain critically understudied. Motivated by these findings, we propose the Agentic Context Bank, a strong inference-time baseline that structures a user's visual context into a self-refining memory bank and employs query-adaptive evidence selection. Our baseline approach consistently improves over standard context prompting regimes across tasks and evaluated backbones, demonstrating a practical path towards future personalized LMMs.\ifarxiv\footnote{Project webpage: \url{https://vision.cs.utexas.edu/projects/PersonalVCL}.}\fi
\end{abstract}


\section{Introduction}
\begin{figure}[t]
\centering
\includegraphics[width=\linewidth]{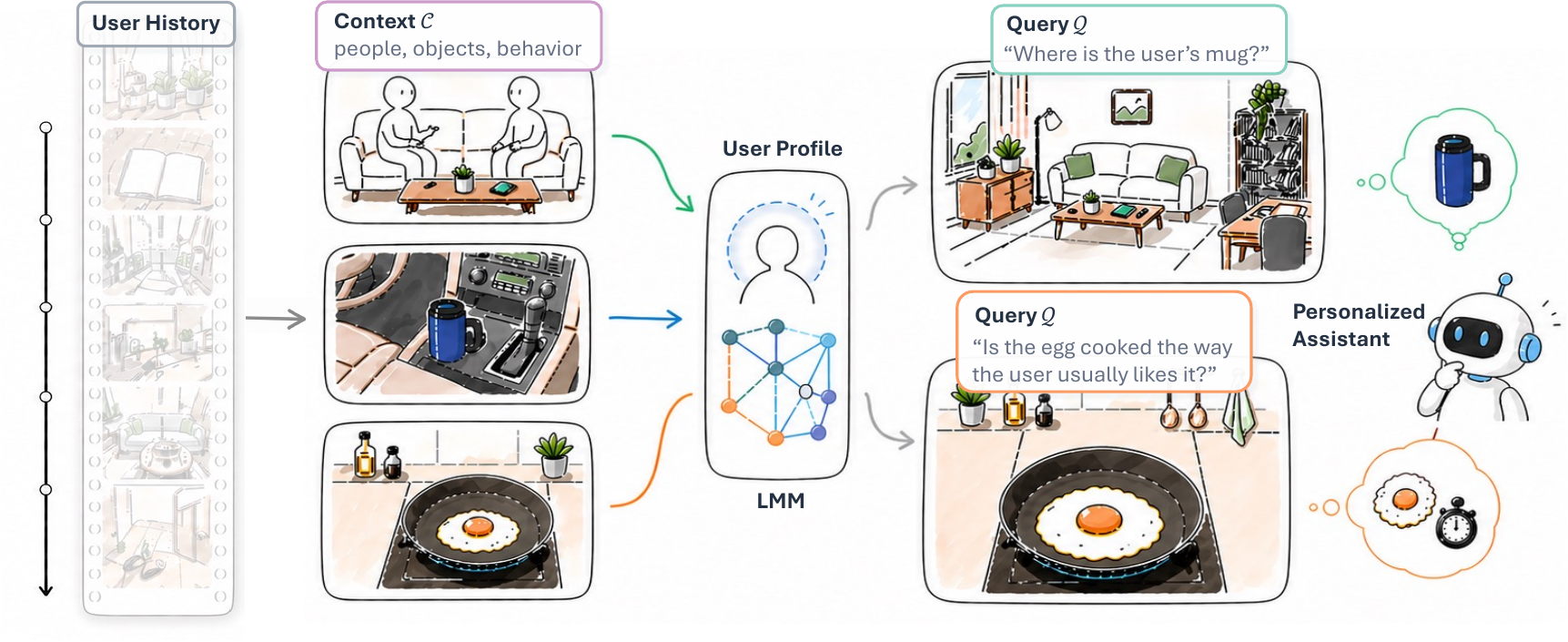}
\caption{
\SX{Personal Visual Context Learning. Continuous egocentric capture from wearable devices presents a user's unique visual history, supplying personal context absent from standard model pre-training. Personal VCL investigates whether an LMM can effectively leverage this context to resolve user-specific queries, ranging from locating a personal object to comparing a current action against the user's past behaviors.}
}
\label{fig:teaser}
\end{figure}

The immense capabilities of Large Multimodal Models (LMMs)~\cite{gpt4, gemini, llava, qwen3vl, internvl35} are undeniable. Looking ahead, wearable technologies like smart glasses~\cite{projectaria, ariadata, vinci} will embed these models into the continuous first-person visual stream of an individual user. This continuous egocentric perception~\cite{ego4d, egoexo, epickitchens, egoexolearn, egolife} is the critical catalyst for turning generalized AI into dedicated personal assistants. We envision a future as depicted in Fig.~\ref{fig:teaser}. Through daily observation, the model constructs an entirely unique visual profile about the user: it maps their social circle, identifies personal items, and knows exactly how they like their fried eggs cooked. Equipped with this visual memory, AI assistants can answer personalized queries and provide active guidance for the user, such as pinpointing where they left their water bottle or alerting if the egg is getting overcooked compared to their usual standard.

This vision fundamentally rests on the broader challenge of model personalization~\cite{llmpersosurvey}. In the text domain, the established mechanism for personalizing LLMs~\cite{lamp, longlamp, pearl, ropg, lampqa} relies on retrieving relevant written information about the user and prepending it to the prompt to guide generation. In this paper, we explore the direct visual analog of this paradigm. 
\SX{As illustrated in Fig.~\ref{fig:teaser}, a user’s visual history consists of the long, continuous egocentric stream captured over time, from which relevant images or clips are drawn to form a \emph{personal visual context}. 
When a user poses a specific question, this personal context is supplied directly alongside the query. A critical question then emerges: can today's LMMs effectively reason over this personal visual evidence to resolve user-specific queries?} We formalize this capability as Personal Visual Context Learning (Personal VCL).

\SX{Our formulation centers around \emph{context utilization}: the ability of an LMM to reason over visual context once it has been supplied in the prompt. This focus distinguishes Personal VCL from three neighboring lines of work.
First, existing LMM personalization works enroll static user-specific concepts (like a specific pet or face) from a few reference images~\cite{palavra} via a dedicated training~\cite{myvlm, yollava, mcllava, pvit, plvlm, unictokens, repic, concepttree, pvchat, mmpb} or storage mechanism~\cite{rap, persotools, onlinepvlm, tame, pearlvideo}, but cannot address inference-time reasoning over complex, dynamic personal context that unfolds across visual observations.
Second, while visual in-context learning~\cite{videoicl, viola, demoicl, vlicl, elicitingicl, manyshoticl, llavaicl} shares our inference-time setting, its use of visual demonstrations serves to elicit task formatting from pre-trained knowledge, not to supply novel, private knowledge about a particular user.
Finally, long-form~\cite{videomme, longvideobench, mvbench, mlvu} and egocentric~\cite{ego4d, egovqa, egotaskqa, groundvqa, egoschema, egolife, hourvideo} video question answering (VQA) target \emph{context acquisition}: how to effectively search a long visual history to identify query-relevant evidence. Personal VCL targets the orthogonal subsequent stage: assuming relevant context has been mined from history, it investigates how to best leverage this visual context for personalized queries.}

The proposed Personal VCL therefore defines a new capability axis for LMMs: prompt-time reasoning over private, fine-grained, and temporally accumulated visual evidence. The fundamental challenge is that the relevant knowledge is neither part of the model's (pre-trained) world knowledge nor contained entirely in the query. 
\SX{It is carried by the user's personal visual context (mined from the broader visual history)}: the faces they know, the objects they own, the routines they repeat, and the small deviations that matter only relative to their own past.
Advancing this capability could enable personal assistants that reason against an individual's own history, from detecting deviations during physical rehabilitation to assessing skill progression in sports, cooking, or crafts.

To chart a course along this path, we introduce Personal-VCL-Bench, a benchmark organized around the three core axes of a user's visual world: persons, objects, and behaviors. It culminates in \codex{EgoWearer identification}, a capstone challenge where the model must recognize the user behind a new egocentric video by aggregating subtle, persistent cues from their prior history: who they interact with, what they own, and how they navigate routine activities. Benchmarking frontier LMMs, we reveal a substantial Personal VCL gap: raw visual context is not reliably exploited, and scaling up the volume of context often fails to improve performance.

\SX{Motivated by this diagnosis, we propose the Agentic Context Bank as a strong inference-time baseline for Personal VCL. Our framework transforms raw visual context into a structured, evidence-linked memory bank and employs query-adaptive selection to inspect only the necessary visual evidence for a given query. Its consistent gains over standard language- and visual-context prompting, across tasks and LMM backbones, underscore context utilization as the central pillar of Personal VCL.}
Together, our formulation, benchmark, diagnosis, and baseline approach establish Personal VCL as a concrete capability 
for building LMMs that can adapt to individuals through visual experience---essential for the next frontier in wearable computing.


\section{Related Work}
\textbf{Personalization.} The trajectory of model personalization begins in the LLM domain, where the user is represented as a body of text~\cite{llmpersosurvey}. Retrieval-based approaches~\cite{lamp, longlamp, pearl, ropg, lampqa} leave weights frozen and pull from the user's written profile on demand, while parameter-based approaches~\cite{oppu, profilepeft, perpcs} absorb that profile into per-user adapters. The visual counterpart to this evolution is concept-centric: a user-specific entity, such as a pet, a face, or an owned object, is enrolled from a few reference images and bound to a dedicated representation, learned per concept~\cite{palavra, myvlm, yollava, mcllava, pvit, plvlm, unictokens, repic, concepttree, pvchat, mmpb}, stored in a retrieval database~\cite{rap}, or matched in frozen feature space~\cite{persotools, onlinepvlm, tame}; recent work further extends the same enrollment paradigm to streaming video~\cite{pearlvideo}.
\SX{Personal VCL moves beyond this concept-enrollment view and investigates personalization as an act of inference-time reasoning over the broader, dynamic visual context of a user's life.}

\textbf{Context learning.} The role of the prompt has expanded beyond simple task specification, increasingly serving as a temporary knowledge base for LLMs during inference~\cite{rag, gemini15}. A capability gap recently identified by CL-bench~\cite{clbench} separates two roles the prompt context can play: it can demonstrate a known task execution pattern, or supply novel knowledge required to answer, with the second proving far harder. The first role is the one that the dominant in-context learning literature has explored, originating in LLMs~\cite{gpt3, minicl, iclsurvey} and carried into the multimodal setting through image~\cite{vlicl, manyshoticl, llavaicl} and video demonstrations~\cite{elicitingicl, videoicl, viola, demoicl}. 
\SX{In those setups, prompt examples act as structural guides that teach the model an input-output mapping rule for tasks reliant on pre-trained knowledge, such as assigning a class label to an image or an action category to a video. In contrast, Personal VCL positions visual context as a private knowledge base. Because the required personal information is inherently absent from any LMM's pre-training, the model must actively reason over the provided visual evidence to discover the novel facts needed to formulate an accurate response.}

\textbf{Long-form and egocentric video understanding.} Personal VCL builds on two closely related lines of video research. In long-form VQA~\cite{videomme, longvideobench, mvbench, mlvu}, the central challenge is context \emph{acquisition}: finding the small set of query-relevant moments from a much longer video. This has driven the development of temporal localization~\cite{momentdetr, tan2d}, planner-observer agents~\cite{agenticvlv, lenswalk, avp, temporalcot, maegoeqa, videoagent}, and memory-augmented stores~\cite{encstore, egoinstructor, esom, eaglelv, lifelongmem, bimba, providellm, rewind, llovi} to construct candidate visual context. Meanwhile, egocentric VQA brings these challenges to first-person capture, featuring tasks like instance search~\cite{ego4d, egovqa}, grounded and task-oriented QA~\cite{egotaskqa, groundvqa}, and multi-day comprehension~\cite{egoschema, egolife, hourvideo}. \SX{Beyond QA, prior studies on wearer identity and privacy~\cite{hoshen2016, thapar2020, egoprivacy} \codex{establish} that first-person footage inherently encodes the camera wearer's identity, an insight we adopt to formulate \codex{EgoWearer identification} in Personal-VCL-Bench. Ultimately, while both VQA domains emphasize acquiring long-horizon visual evidence, Personal VCL isolates the critical subsequent step of context \emph{utilization}:}
given context drawn from the user's history, 
the model 
must use it to respond to a user-specific query---even when the response requires indirect reasoning.


\section{Personal Visual Context Learning}
\label{sec:problem}

This section formalizes the problem of Personal VCL (Sec.~\ref{sec:formulation}) and introduces Personal-VCL-Bench (Sec.~\ref{sec:bench}). We leverage this benchmark to conduct an in-depth empirical investigation into frontier LMM performance (Sec.~\ref{sec:investigation}),  
\SX{and derive a strong baseline approach, the Agentic Context Bank, from these findings (\codex{Sec.~\ref{sec:method}}).}
\subsection{Problem Formulation}
\label{sec:formulation}

Personal VCL assesses whether a model can leverage personal visual history as context to answer user-specific visual queries. Formally, each instance is a pair $(\mathcal{C}, \mathcal{Q})$. The context $\mathcal{C} = (\mathcal{L}_c, \mathcal{V}_c)$ pairs a language declaration $\mathcal{L}_c$ that names what the context shows 
with the corresponding visual material $\mathcal{V}_c$ it refers to. Because personal context accumulates over time, we consider $\mathcal{V}_c$ as a collection of reference observations from the user's history, such as several images of the same person or object, or several clips of the same routine. The query $\mathcal{Q} = (\mathcal{L}_q, \mathcal{V}_q)$ is structured analogously: $\mathcal{L}_q$ poses a question about the user in relation to the image or video $\mathcal{V}_q$. This language query $\mathcal{L}_q$ may correspond to an explicit user request, or to an internal decision point generated by a broader assistant system (e.g., a proactive agent evaluating whether a user's current physical execution deviates from their historical baseline before triggering an automated correction). A generic pretrained LMM $f_\theta$, frozen at inference time, is tasked with responding to the query $\mathcal{Q}$ based on user context $\mathcal{C}$.

\subsection{Personal-VCL-Bench}
\label{sec:bench}
What constitutes our personal visual world? It is fundamentally shaped by the specific people we encounter, the unique objects we interact with, and the distinct activities we perform. Grounded in this philosophy, we introduce Personal-VCL-Bench, structured precisely along these three fundamental axes of personal visual knowledge: persons, objects, and behavior, with \codex{EgoWearer identification} serving as an integrative task over all three.

\begin{figure}[!t]
\centering
\includegraphics[width=\linewidth]{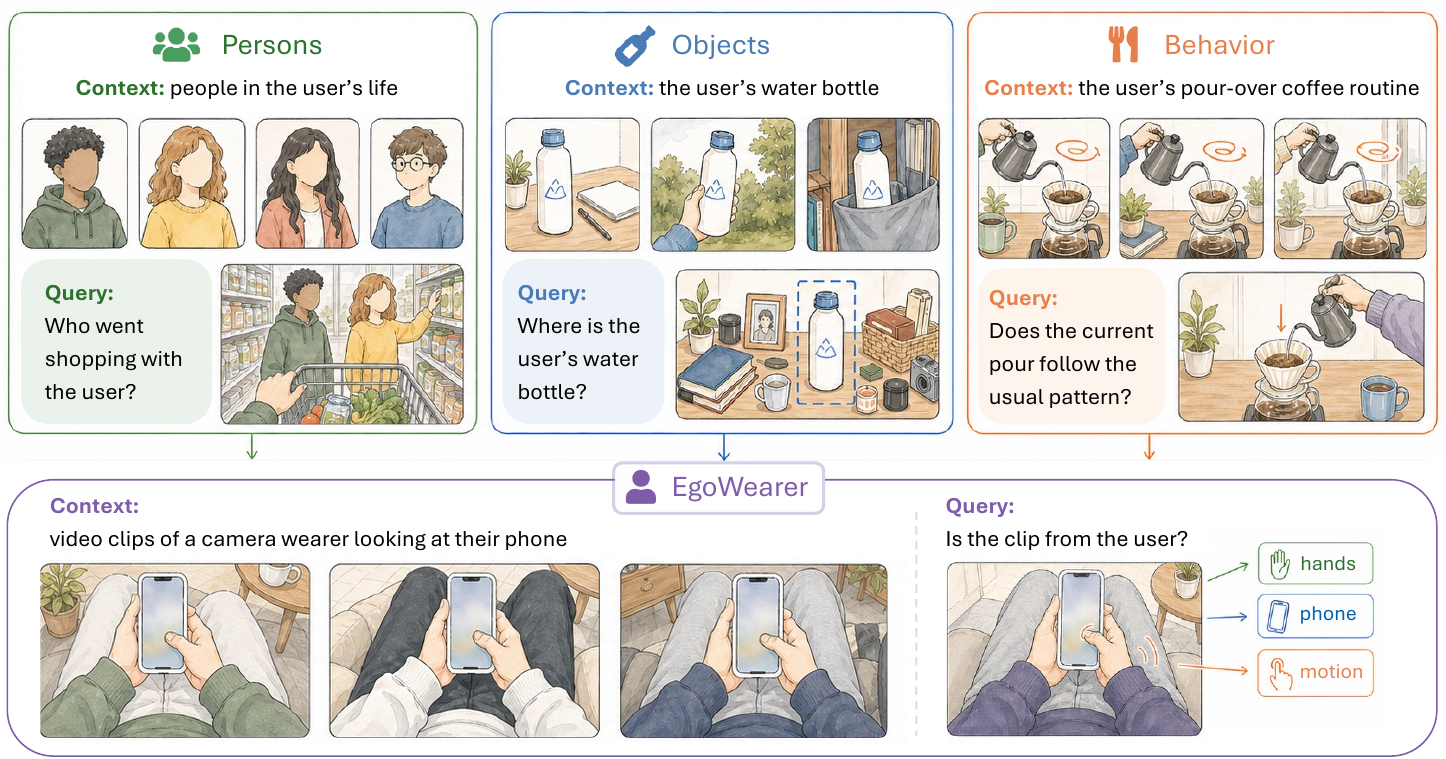}
\caption{
\SX{We propose Personal-VCL-Bench to evaluate how LMMs use personal visual context across three axes: persons, objects, and behavior. Each panel shows a representative context--query example. The \codex{EgoWearer Identification} task serves as the benchmark capstone, where the model must decide whether a query clip belongs to the same camera wearer by connecting indirect cues across axes, such as hands, the phone, and the \codex{swiping} motion in this example.}}
\label{fig:benchmark}
\end{figure}

We instantiate these axes using three egocentric datasets (EgoLife~\cite{egolife}, Ego4D~\cite{ego4d}, and CaptainCook4D~\cite{captaincook4d}) that capture the day-to-day perspectives of individual users across multiple scenes. We manually design the task semantics, formulate user-specific queries, and curate the visual contexts needed to answer them. \SX{This yields \codex{2,255} clean context-query instances spanning 7 tasks; each query is paired with 4.25 supplied context images/clips on average.} 
Fig.~\ref{fig:benchmark} illustrates concrete benchmark instances across all axes, showing how personal context is paired with a user-specific query. See Supp.~\ref{sec:supp-bench} for full benchmark details.


\textbf{Persons.} We posit that a personal assistant must reliably navigate the immediate social circle of a user. To evaluate this capability, we instantiate our persons axis on EgoLife~\cite{egolife}, a multi-day egocentric recording of six co-living participants, via two progressive tasks. The first is single identity recognition, acting as an essential proof of concept for matching a target person using a provided reference gallery. \SX{The second task, derived from the RelationQA set of EgoLifeQA benchmark~\cite{egolife}, requires the model to resolve multiple people’s identities from context and 
then map those specific identities onto the query video to successfully interpret the social dynamics occurring between them.}

\textbf{Objects.} The true utility of a visual assistant lies not in recognizing a generic category like a bottle, but in identifying the exact bottle owned by the user. Queries along this axis fundamentally demand instance-level rather than category-level reasoning. 
We ground this axis by repurposing the Ego4D-VQ benchmark~\cite{ego4d}
\SX{into two complementary tasks: object identity recognition, where the model decides whether the context object appears in a query video, and personalized object detection, where the model localizes the context object instance in a query frame with a bounding box.}

\textbf{Behavior.} Human routines are highly individualized, characterized by distinct execution patterns, tempos, and degrees of dexterity. The objective along this axis is to evaluate an action not against a universal standard, but against the user's own baseline. \SX{For example, a wearable assistant might track changes in a patient’s dexterity during stroke rehabilitation, or compare a cook’s current knife technique against their usual safe motion.}
To achieve this, we adapt CaptainCook4D~\cite{captaincook4d} to formulate two tasks. Given a few reference videos of a user's standard execution as context, a model is tasked with analyzing a new query video of the same user to either identify procedural mistakes or pinpoint fine-grained behavioral consistencies and deviations.

\textbf{\codex{EgoWearer identification}.} A personal assistant is fundamentally anchored to the identity of its user; its memory and actions are only valid if it can maintain continuity over who is wearing the device. We propose EgoWearer identification to test this memory-binding capability: given visual context from one user's history, the model needs to decide whether a new egocentric clip belongs to the same wearer. \SX{Crucially, since the first-person viewpoint inherently conceals the wearer, the model cannot rely on direct observation. It must instead construct a coherent profile of the ``unseen'' user by piecing together indirect visual evidence, such as the objects they repeatedly use, or the distinctive ways they perform routine actions. By demanding this complex deductive reasoning, \codex{EgoWearer identification} stands as the capstone of Personal-VCL-Bench, providing a stress test for fine-grained personalization and a critical prerequisite for continuous authentication in real-world assistants.}

\subsection{Diagnosing the Personal VCL Gap}
\label{sec:investigation}

\begin{table}[t]
\centering
\bencheval
\newcommand{\visup}[2]{#1\rlap{\,\textsuperscript{\textcolor{green!45!black}{$\boldsymbol{\uparrow}$}}}}
\newcommand{\visdown}[2]{#1\rlap{\,\textsuperscript{\textcolor{red!65!black}{$\boldsymbol{\downarrow}$}}}}
\newcommand{\vistie}[2]{#1\rlap{\,\textsuperscript{\textcolor{black!55}{$\boldsymbol{\sim}$}}}}
\caption{Diagnosing frontier LMMs on Personal-VCL-Bench across \codex{five} context regimes (per-task accuracy, \%). \textbf{Bold} = best within a model's five rows. $k{=}1$ denotes using a single context item, while $k{=}k_{\max}$ denotes using all available context items for each query. In visual-context rows, cell shading indicates whether visual context improves, hurts, or ties against the matched language-context row. In visual-ctx ($k_{\max}$) cells, superscript markers indicate whether increasing visual context from $k{=}1$ to $k_{\max}$ improves, drops, or ties. \SX{Taken together, the results reveal that current LMMs struggle to capitalize on the primary assets of Personal VCL: utilization of raw visual evidence and integration of multiple context observations.} Full results are in Tab.~\ref{tab:investigation-full} of Supp.~\ref{sec:supp-diagnostic}.}
\label{tab:investigation}
\begin{tabular}{l*{6}{N}E}
\benchheader
Random guess & 50.00 & 25.00 & 50.00 & -- & 50.00 & 25.00 & 50.00 \\
\benchmidrule
\multicolumn{8}{c}{Qwen3-VL-8B~\cite{qwen3vl}} \\
No-context              & 50.50 & 36.00 & 49.06 & 68.12 & 49.46 & 19.39 & 49.83 \\
Language-ctx ($k{=}1$)    & 51.75 & 29.60 & 65.59 & 76.09 & \textbf{57.57} & \textbf{24.49} & 52.57 \\
Language-ctx ($k_{\max}$) & 62.75 & \textbf{48.80} & 67.25 & \textbf{77.54} & 54.86 & 22.45 & \textbf{54.53} \\
Visual-ctx ($k{=}1$)      & \cellcolor{cellgreen}62.75 & \cellcolor{cellgreen}46.40 & \cellcolor{cellgreen}70.99 & \cellcolor{cellgreen}\textbf{77.54} & \cellcolor{cellgray}\textbf{57.57} & \cellcolor{cellred}21.43 & \cellcolor{cellred}50.00 \\
Visual-ctx ($k_{\max}$)   & \cellcolor{cellgreen}\visup{\textbf{85.25}}{} & \cellcolor{cellred}\vistie{46.40}{} & \cellcolor{cellgreen}\visup{\textbf{73.67}}{} & \cellcolor{cellgray}\vistie{76.81}{} & \cellcolor{cellgray}\visdown{55.14}{} & \cellcolor{cellgray}\visup{22.45}{} & \cellcolor{cellred}\vistie{50.00}{} \\
\benchmidrule
\multicolumn{8}{c}{Gemma-4-31B~\cite{gemma4}} \\
No-context              & 49.25 & 22.40 & 51.08 & 57.97 & 52.43 & 29.59 & 50.13 \\
Language-ctx ($k{=}1$)    & 57.50 & 20.80 & 78.33 & \textbf{72.46} & 54.86 & 33.67 & 54.52 \\
Language-ctx ($k_{\max}$) & \textbf{78.00} & \textbf{49.60} & \textbf{78.75} & 71.74 & 58.11 & \textbf{34.69} & 52.99 \\
Visual-ctx ($k{=}1$)      & \cellcolor{cellgreen}65.50 & \cellcolor{cellgreen}30.40 & \cellcolor{cellred}73.26 & \cellcolor{cellred}63.77 & \cellcolor{cellgreen}\textbf{59.46} & \cellcolor{cellred}32.65 & \cellcolor{cellgreen}\textbf{56.66} \\
Visual-ctx ($k_{\max}$)   & \cellcolor{cellred}\visup{75.75}{} & \cellcolor{cellred}\visup{46.40}{} & \cellcolor{cellred}\visup{75.34}{} & \cellcolor{cellred}\visup{65.22}{} & \cellcolor{cellgreen}\vistie{\textbf{59.46}}{} & \cellcolor{cellred}\visdown{27.55}{} & \cellcolor{cellgreen}\visdown{55.36}{} \\
\benchmidrule
\multicolumn{8}{c}{Gemini-3-Flash~\cite{gemini3flash}} \\
No-context              & 55.75 & 42.40 & 49.32 & 64.49 & 57.84 & 37.76 & 49.47 \\
Language-ctx ($k{=}1$)    & 69.25 & 37.60 & 80.55 & 79.71 & 55.41 & 45.92 & 53.99 \\
Language-ctx ($k_{\max}$) & 81.75 & \textbf{59.20} & 82.35 & 78.99 & 54.59 & 48.98 & 55.26 \\
Visual-ctx ($k{=}1$)      & \cellcolor{cellred}67.50 & \cellcolor{cellgreen}44.00 & \cellcolor{cellgreen}82.21 & \cellcolor{cellgreen}\textbf{85.51} & \cellcolor{cellgreen}\textbf{68.92} & \cellcolor{cellgreen}47.96 & \cellcolor{cellred}51.99 \\
Visual-ctx ($k_{\max}$)   & \cellcolor{cellgreen}\visup{\textbf{88.00}}{} & \cellcolor{cellred}\visup{54.40}{} & \cellcolor{cellgray}\vistie{\textbf{82.47}}{} & \cellcolor{cellgreen}\visdown{82.61}{} & \cellcolor{cellgreen}\visdown{67.03}{} & \cellcolor{cellgreen}\visup{\textbf{61.22}}{} & \cellcolor{cellgray}\visup{\textbf{55.74}}{} \\
\benchbottomrule
\end{tabular}
\end{table}

To systematically dissect the context utilization problem, we structure our investigation around two core questions: how should personal visual context be represented, and how effectively do models integrate multiple pieces of evidence?
For the first question, we evaluate a visual-context regime that provides the reference material $\mathcal{C}$ as raw pixels against a language-context regime that provides an LMM-generated textual description of that visual content, allowing us to assess whether native visual data offers advantages over textual proxies. A no-context baseline is also established using only the query $\mathcal{Q}$ to quantify the model's reliance on generic pre-trained knowledge.
To address the second question, we compare model performance when provided with a single context ($k{=}1$) versus all available context items ($k{=}k_{\max}$) across both context regimes.
We evaluate seven frontier LMMs on Personal-VCL-Bench. \codex{Tab.~\ref{tab:investigation}} reports a representative subset, with the full results provided in Supp.~\ref{sec:supp-diagnostic}. Two key observations emerge:

\textbf{The modality paradox.} Human cognition effortlessly processes raw visual snapshots; we recall the exact appearance of a familiar face or a personal object without needing to first translate those memories into words. However, our results indicate that current LMMs struggle to natively emulate this capability. In a great number of our evaluated cases (red/gray cells in \codex{Tab.~\ref{tab:investigation}}), providing the model with a language context, which is inherently a lossy semantic compression of the full visual signal, yields performance that is on par with or even superior to using the raw pixels. This paradox reveals a substantial deficiency in VCL compared to its textual counterpart. We hypothesize that this limitation stems from the foundational training paradigms of frontier LMMs, which remain overwhelmingly optimized for language processing rather than genuine, native visual reasoning.

\textbf{The scaling paradox.} A fundamental assumption of in-context learning is that access to more evidence yields better predictions. Yet, in Personal VCL, LMMs contradict this expectation. When comparing a single context item against all available context (1-vs-all; superscript arrows in \codex{Tab.~\ref{tab:investigation}}), the addition of relevant historical data frequently fails to improve accuracy. This reveals a critical inability of current models to synthesize multiple visual observations and extract consistent personal patterns. Since real-world visual histories grow continuously over time, overcoming this scaling bottleneck is imperative for the development of robust personal agents.

\subsection{Agentic Context Bank}
\label{sec:method}

\SX{The diagnostic results suggest that Personal VCL is far from being solved by simply appending and concatenating visual context into a prompt. Instead, they point to a key context utilization problem: how should the supplied visual context be best organized and presented to the LMM? Driven by the two observations above, we introduce the Agentic Context Bank, a strong model-agnostic inference-time baseline for Personal VCL. The framework operates in two stages. Stage I constructs a structured bank by converting disjoint context items into a coherent, self-refining memory (Sec.~\ref{sec:bank-construction}). Stage II applies query-adaptive evidence selection. At inference time, the model first surveys a lightweight text view of the bank, choosing to load the actual visual evidence only for the entries required to resolve the active query (Sec.~\ref{sec:evidence-selection}). The full procedure is outlined in Algorithm~\ref{alg:bank}.}

\begin{algorithm}[t]
\caption{Agentic Context Bank}
\label{alg:bank}
\KwIn{visual context $\mathcal{V}_c = \{v_1, \dots, v_k\}$, query $\mathcal{Q} = (\mathcal{L}_q, \mathcal{V}_q)$, LMM $f_\theta$}
\KwOut{answer $A$}
\tcp{Stage I: Structured Bank Construction (query-agnostic; Sec.~\ref{sec:bank-construction})}
$\mathcal{B} \leftarrow \emptyset$\;
\For{$v_i \in \mathcal{V}_c$}{
  $\mathcal{M}_i \leftarrow \textsc{Extract}(v_i)$ \tcp*[r]{candidate memory entries $(\tau, d, e)$}
  \ForEach{$(\tau, d, e) \in \mathcal{M}_i$}{
    $op \leftarrow \textsc{Merge}(\mathcal{B},\, (\tau, d, e))$ \tcp*[r]{$op \in \{\textsc{add}, \textsc{confirm}, \textsc{revise}, \textsc{retract}\}$}
    apply $op$ to $\mathcal{B}$\;
  }
}
\tcp{Stage II: Query-Adaptive Evidence Selection (query-specific; Sec.~\ref{sec:evidence-selection})}
$T_{\mathcal{B}} \leftarrow \textsc{TextView}(\mathcal{B})$ \tcp*[r]{memory descriptors with entry IDs}
$y \leftarrow f_\theta\big(T_{\mathcal{B}},\, \mathcal{L}_q,\, \mathcal{V}_q\big)$ \tcp*[r]{Call 1: text triage}
\uIf{$y=(\mathrm{request}, \mathcal{I})$}{
  $H_{\mathcal{I}} \leftarrow \textsc{HybridView}(\mathcal{B}, \mathcal{I})$ \tcp*[r]{inline evidence only for selected IDs $\mathcal{I}$}
  $A \leftarrow f_\theta\big(H_{\mathcal{I}},\, \mathcal{L}_q,\, \mathcal{V}_q\big)$ \tcp*[r]{Call 2: selective visual verification}
}
\Else{
  $A \leftarrow y.\mathrm{answer}$\;
}
\Return $A$\;
\end{algorithm}

\subsubsection{Stage I: Structured Bank Construction}
\label{sec:bank-construction}

The goal of stage I is to convert a set of raw context observations into a structured memory that can grow with the user. Rather than treating the $k$ context items in $\mathcal{V}_c$ as independent prompt examples, we view them as compounding evidence about the same underlying individual. The bank $\mathcal{B}$ is therefore represented as a set of entries $(\tau, d, e)$. Here, $\tau$ denotes the memory type, \SX{\codex{taking values in} $\{\textsc{appearance}, \textsc{owned\_objects}, \textsc{behavior}\}$ and broadly aligned with the three benchmark axes.}
$d$ is a natural-language memory descriptor grounded by the corresponding visual evidence $e$. Depending on the memory type, $e$ takes two forms: a single supporting frame for $\textsc{appearance}$ and $\textsc{owned\_objects}$ to capture their static properties, or a video span for $\textsc{behavior}$ to preserve the temporal structure. 

The bank is updated sequentially as new context items are processed. \textsc{Extract} converts each item into a set of candidate memory entries $\mathcal{M}_i$; \textsc{Merge} compares those candidates against the current bank and applies one of four updates. \textsc{Add} creates a new entry, \textsc{confirm} accumulates support for an existing entry, \textsc{revise} refines the descriptor when the new observation makes it more precise, and \textsc{retract} 
\SX{removes an entry from the bank when later extracted cues suggest it is unreliable, such as a transient cue or an earlier visual misread.} In this way, Stage I turns an expanding visual history into a compact, evidence-linked memory rather than a naive concatenation of prompt examples.

\subsubsection{Stage II: Query-Adaptive Evidence Selection}
\label{sec:evidence-selection}

With the memory bank $\mathcal{B}$ established, the subsequent challenge is to find the right information for a given query. Because the bank aggregates rich appearance cues, personal objects, and behavioral routines from multiple clips, supplying a model with the entirety of this raw visual data is highly inefficient. At the same time, relying solely on the textual view of the bank is inherently lossy and sacrifices the precise visual details required for fine-grained personalization. To resolve this, we use each natural-language memory descriptor as a textual index into stored visual evidence. The model first surveys the text view $T_{\mathcal{B}}$ to identify a selected set of entry IDs $\mathcal{I}$ relevant to the current query. It then receives a hybrid view $H_{\mathcal{I}}$ that inlines visual evidence only for those selected entries. This process yields a query-tailored memory view.

We implement this as a two-step inference process. First, during an initial text triage, the LMM is presented with $T_{\mathcal{B}}$, listing $\tau$, $d$, and a stable ID for each entry alongside the query $\mathcal{L}_q$ and $\mathcal{V}_q$. Based on this lightweight summary, the model either answers the query directly or issues a tool call requesting the visual evidence $e$ for specific entry IDs. Second, we execute selective visual verification. If a tool call was issued, the bank is re-rendered as $H_{\mathcal{I}}$, which inlines the supporting frames or video spans strictly for the requested IDs while leaving the rest of the bank as text. The model then produces its final answer based on this hybrid context. By treating evidence selection as an explicit reasoning decision, our design ensures the model only loads visual evidence into the heavier context window when necessary. 

\SX{Together, the two stages turn the diagnostic findings in Sec.~\ref{sec:investigation} into a concrete Personal VCL baseline. Stage I tackles the scaling paradox, by shifting the unit of context from isolated clips to persistent, evidence-linked personal cues. Stage II addresses the modality paradox, by replacing passive visual concatenation with query-conditioned selection, which preserves the benefit of textual organization while retaining access to the supporting visual evidence. By establishing this structured approach, we aim to provide a principled starting point for subsequent research in Personal VCL.}

\section{Experiments}
\label{sec:experiments}

\subsection{Setup}
\label{sec:setup}

\textbf{Evaluation.}
We evaluate the Agentic Context Bank as a strong inference-time baseline for Personal VCL across the full Personal-VCL-Bench suite using Gemma-4-31B~\cite{gemma4}, a representative strong \codex{open-weight} LMM. Since EgoWearer identification integrates all three axes and places the strongest demand on context utilization, we further evaluate this task across Gemma-4-31B~\cite{gemma4}, Gemini-3-Flash~\cite{gemini3flash}, and GPT-5.4-mini~\cite{gpt54} to test whether the baseline generalizes across LMM backbones.
Implementation details are provided in Supp.~\ref{sec:supp-implementation}.

\textbf{Baselines.} We compare against standard context prompting baselines.
The language-context regime converts each item in $\mathcal{V}_c$ into a textual description and feeds only the descriptions to the LMM, a verbalize-and-answer pipeline widely used in long-form video understanding~\cite{llovi, lifelongmem, encstore}. The visual-context regime concatenates the raw visual tokens of all items in $\mathcal{V}_c$ into the model prompt, the dominant setup in multi-image and many-shot visual in-context learning~\cite{vlicl, manyshoticl, llavaicl}. \SX{These comparisons place the Agentic Context Bank against the main existing choices for visual context utilization: no context, text summaries, or flat visual concatenation.}

\subsection{Results}
\label{sec:main-result}
\textbf{Task-wise results.} Tab.~\ref{tab:gemma-taskwise} evaluates the Agentic Context Bank across the full Personal-VCL-Bench suite. Across persons, objects, behavior, and EgoWearer identification, the bank improves over language-context and visual-context prompting baselines. The gains are largest on tasks that require aggregating or comparing personal evidence, such as EgoWearer identification. These results support our central claim that visual personalization requires not only access to relevant context, but also a better way to organize and use it.

\begin{table}[!t]
\centering
\bencheval
\caption{Task-wise results of the Agentic Context Bank on Gemma-4-31B~\cite{gemma4} across Personal-VCL-Bench (per-task accuracy, \%). \SX{The comparisons span the main ways current LMMs are given context: No-context for query-only prompting, Language-ctx for textual descriptions of each context item~\cite{llovi, lifelongmem, encstore}, and Visual-ctx for direct prompting with the raw context images or clips~\cite{vlicl, manyshoticl, llavaicl}}. Absolute gain is measured over the stronger of Language-ctx and Visual-ctx at $k_{\max}$.}
\label{tab:gemma-taskwise}
\begin{tabular}{l*{6}{N}E}
\benchheader
No-context
  & 49.25 & 22.40 & 51.08 & 57.97 & 52.43 & 29.59 & 50.13 \\
Language-ctx ($k_{\max}$)
  & 78.00 & 49.60 & 78.75 & 71.74 & 58.11 & 34.69 & 52.99 \\
Visual-ctx ($k_{\max}$)
  & 75.75 & 46.40 & 75.34 & 65.22 & 59.46 & 27.55 & 55.36 \\
\benchmidrule
\rowcolor{cellgreen}
Ours
  & \textbf{83.25} & \textbf{51.20} & \textbf{82.52} & \textbf{73.19} & \textbf{66.22} & \textbf{35.71} & \textbf{61.60} \\
\rowcolor{cellgreen}
(Gain)
  & \textcolor{green!45!black}{\footnotesize(+5.25)}
  & \textcolor{green!45!black}{\footnotesize(+1.60)}
  & \textcolor{green!45!black}{\footnotesize(+3.77)}
  & \textcolor{green!45!black}{\footnotesize(+1.45)}
  & \textcolor{green!45!black}{\footnotesize(+6.76)}
  & \textcolor{green!45!black}{\footnotesize(+1.02)}
  & \textcolor{green!45!black}{\footnotesize(+6.24)} \\
\benchbottomrule
\end{tabular}
\end{table}

\begin{table}[!t]
\centering
\bencheval
\caption{\codex{EgoWearer Identification} as an integrated Personal VCL test (accuracy, \%). We evaluate the Agentic Context Bank across three LMM backbones and ablate stage-II evidence selection by comparing adaptive evidence selection against descriptors-only and all-evidence variants. The consistent gains indicate that Personal VCL requires structured and selective access to visual context.}
\label{tab:main}
\begin{tabular}{ll*{3}{>{\centering\arraybackslash}p{2.7cm}}}
\toprule
& Method & GPT-5.4-mini~\cite{gpt54} & Gemma-4-31B~\cite{gemma4} & Gemini-3-Flash~\cite{gemini3flash} \\
\midrule
\multirow{3}{*}{Baselines}
  & No-context             & 49.83 & 50.13 & 49.47 \\
  & Language-ctx ($k_{\max}$)  & 49.86 & 52.99 & 55.26 \\
  & Visual-ctx ($k_{\max}$) & 51.47 & 55.36 & 55.74 \\
\specialrule{\lightrulewidth}{0pt}{0pt}
\multirow{3}{*}{Ours}
  & Descriptors only       & 50.69 & 51.97 & 55.28 \\
  & All evidence       & 51.21 & 57.29 & 67.86 \\
  & \cellcolor{cellgreen}Adaptive evidence
  & \cellcolor{cellgreen}\textbf{53.77}
  & \cellcolor{cellgreen}\textbf{61.60}
  & \cellcolor{cellgreen}\textbf{72.82} \\
\specialrule{\heavyrulewidth}{0pt}{0pt}
\end{tabular}
\end{table}


\textbf{Cross-backbone analysis.} Tab.~\ref{tab:main} studies EgoWearer identification, the benchmark capstone that requires integrating appearance, owned-objects, and behavioral cues. The full Agentic Context Bank consistently outperforms standard context prompting across all evaluated backbones, with the largest gain reaching 17.1\% over the strongest standard baseline on Gemini-3-Flash~\cite{gemini3flash}. These results indicate that the benefit of structured, query-adaptive memory is not tied to a single model family.

\textbf{Stage-II ablation.} The lower block of Tab.~\ref{tab:main} ablates the design of \codex{query-adaptive} evidence selection. We compare against two ways of querying the same bank. The first uses only the text view $T_\mathcal{B}$, exposing the structured memory descriptors without any stored visual evidence. The second exposes visual evidence for all bank entries, corresponding to an exhaustive version of the hybrid view. Both variants underperform adaptive selection, showing that the gain comes not only from building a better memory, but also from using it adaptively at query time. See Supp.~\ref{sec:supp-additional} for stage-I ablation and additional analysis.

\begin{figure}[t]
\centering
\includegraphics[width=\linewidth]{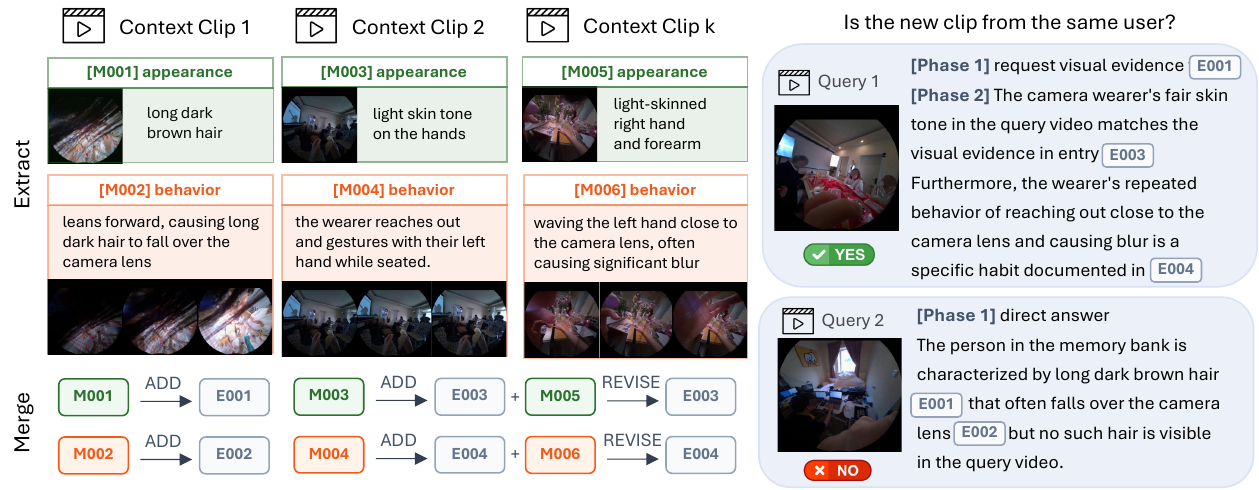}
\caption{
\SX{Qualitative example of the full Agentic Context Bank pipeline. The left side shows stage-I bank construction: from each context clip, the model extracts candidate memory entries (e.g., M001, M002) describing visually-grounded appearance or behavior cues. Through a merging step, these candidates are consolidated into stable evidence-linked bank entries (e.g., E001, E002), which preserve the supporting visual evidence for later inspection. The right side shows stage-II query-time use. Given a new query clip, the model first reasons over the text view of the bank and either requests selected evidence entries for visual verification or answers directly. In Query 1, the model requests relevant evidence and confirms the same wearer; in Query 2, the text descriptors already contradict the query clip, allowing a direct negative decision. We encourage readers to view the Supp. video for full temporal context.}
}
\label{fig:qual}
\end{figure}

\textbf{Qualitative examples.} Fig.~\ref{fig:qual} provides an example of the full bank lifecycle. From several context clips, the model extracts typed memory entries such as hand appearance and motion patterns. The merge process then updates the bank by adding new cues and revising overlapping ones, producing a compact memory rather than a flat list of clips. At inference time, the same bank supports different reasoning paths: the first query triggers targeted visual verification of relevant entries, while the second can be answered directly from the textual memory because the visible wearer contradicts stored identity cues.

\textbf{Limitations.} \SX{This work studies context utilization in isolation, assuming relevant moments are already retrieved from a continuous personal history. Because prior research has dedicated significant attention to the context acquisition problem, we enforce this specific boundary to demonstrate that frontier LMMs still fail even under this simplified condition (Sec.~\ref{sec:investigation}). Extending our framework to jointly evaluate retrieval and reasoning is an important next step. Furthermore, our Agentic Context Bank is intended as a diagnostic baseline for this context utilization problem; its improvements identify useful ingredients, while its remaining errors (see Supp.~\ref{sec:supp-additional} for failure cases) highlight a clear need for continued innovation in this domain.}

\section{Conclusion}
\label{sec:conclusion}
The path from generic LMMs to genuinely personal assistants runs through a capability that pre-training cannot supply by construction: reasoning from a user's own visual history, over knowledge that is unique to that user. We introduce Personal VCL to formalize this capability, alongside Personal-VCL-Bench to systematically evaluate LMM performance.
Our investigation diagnoses a severe context utilization gap in frontier LMMs, characterized by a modality paradox (an over-reliance on the lossy text modality) and a scaling paradox (an inability to reliably aggregate expanding visual context). To address both, we propose the Agentic Context Bank. This inference-time framework aggregates context items into a structured, self-refining memory that is consulted via query-adaptive evidence selection, greatly narrowing the performance gap.

Personal VCL is, we believe, one of the defining capabilities on the trajectory towards personalized AI systems. We hope the problem formulation, benchmark, diagnostic findings, and \SX{baseline} method this paper offers are a useful starting point for that effort. \SX{To support the broader research community, we will publicly release all aspects of Personal-VCL-Bench and our baseline implementations.}

\newpage
\bibliographystyle{plain}
\bibliography{ref}

\defaultfloatspacing
\makeatletter
\setlength{\@fptop}{0pt}
\makeatother
\newpage
\appendix

\section*{Supplementary Material}
\vspace{-0.4em}
\noindent\textbf{Contents}
\vspace{0.4em}

\begingroup
\hypersetup{linkcolor=citecolor}
\normalsize
\setlength{\parindent}{0pt}
\newcommand{\suppcontentsline}[3]{%
  \noindent\makebox[1.7em][l]{\textbf{#1}}%
  \hyperref[#2]{#3}\hfill{\hypersetup{linkcolor=citecolor}\hyperref[#2]{\pageref{#2}}}\par
}
\suppcontentsline{A}{sec:supp-bench}{Personal-VCL-Bench}
\suppcontentsline{B}{sec:supp-diagnostic}{Extended Diagnostic Results}
\suppcontentsline{C}{sec:supp-implementation}{Implementation Details}
\suppcontentsline{D}{sec:supp-additional}{Additional Results and Analysis}
\suppcontentsline{E}{sec:supp-impact}{Broader Impacts}
\endgroup
\vspace{0.8em}

\section{Personal-VCL-Bench}
\label{sec:supp-bench}

\subsection{Task summary}
Personal-VCL-Bench is built by repurposing source videos from EgoLife~\cite{egolife}, Ego4D~\cite{ego4d}, and CaptainCook4D~\cite{captaincook4d} into personalized context-query tasks. \codex{The source datasets are used under their original terms}: EgoLife is listed as MIT-licensed in its Hugging Face release, Ego4D is governed by the Ego4D License Agreement, and CaptainCook4D is released under Apache License 2.0.

The benchmark is constructed to isolate a central question: given curated visual evidence from a user's history, can an LMM use that evidence to answer a new user-specific visual query? Towards this end, we design seven tasks spanning the major forms of personal visual knowledge an assistant must acquire: familiar people, personally relevant objects, individualized behavior, and the wearer's own identity. Tab.~\ref{tab:bench} reports the task organization, context-query modality, task size, and the average number of context images or clips available for each query. Our design stresses two core challenges of Personal VCL: fine-grained visual understanding, and the need to aggregate evidence from multiple context observations. Together, these properties make the benchmark a targeted testbed for diagnosing Personal VCL in LMMs.

\begin{table}[!b]
\centering
\bencheval
\caption{Personal-VCL-Bench task summary. Seven tasks span three axes of personal visual knowledge (persons, objects, behavior) plus the \codex{EgoWearer Identification} capstone. $\overline{|\mathcal{V}_c|}$ reports the average number of context images or clips per query. \faImage~= image, \faFilm~= video.}
\label{tab:bench}
\begin{tabular}{lllllcc}
\toprule
Axis & Task ID & Task & Source & $\mathcal{V}_c \rightarrow \mathcal{V}_q$ & $\overline{|\mathcal{V}_c|}$ & \# \\
\midrule
\multirow{2}{*}{\textcolor{axispersons}{\faUsers~Persons}}
  & PerID  & Single-identity recognition & EgoLife~\cite{egolife}            & \faImage~$\rightarrow$ \faImage & 5.00  & 250 \\
  & PerRel & Relationship reasoning       & EgoLife~\cite{egolife}            & \faImage~$\rightarrow$ \faFilm  & 25.00 & 125 \\
\midrule
\multirow{2}{*}{\textcolor{axisobjects}{\faWineBottle~Objects}}
  & ObjID  & Object identity recognition        & Ego4D~\cite{ego4d}                & \faImage~$\rightarrow$ \faFilm  & 1.42  & 660 \\
  & ObjDet & Personalized object detection           & Ego4D~\cite{ego4d}                & \faImage~$\rightarrow$ \faImage & 1.72  & 138 \\
\midrule
\multirow{2}{*}{\textcolor{axisbehavior}{\faUtensils~Behavior}}
  & BehErr & Procedural error detection  & CaptainCook4D~\cite{captaincook4d} & \faFilm~$\rightarrow$ \faFilm  & 1.98  & 370 \\
  & BehQA  & Procedural question answering & CaptainCook4D~\cite{captaincook4d} & \faFilm~$\rightarrow$ \faFilm & 2.40  & 98 \\
\midrule
\textcolor{axisego}{\faUser~EgoWearer}
  & EgoID  & \codex{EgoWearer Identification}    & EgoLife~\cite{egolife}            & \faFilm~$\rightarrow$ \faFilm   & 5.00  & 614 \\
\bottomrule
\end{tabular}
\end{table}

\subsection{Task construction}
\paragraph{Persons.} The persons axis is designed to test whether Personal VCL can support social understanding, beginning with visual identity binding and extending to relationship reasoning. EgoLife~\cite{egolife} provides multi-day egocentric recordings from a shared household. We focus on five housemates observed from the camera wearer (Jake)'s perspective: Alice, Katrina, Lucia, Shure, and Tasha. For each housemate, we manually select clear Day-1 reference images to form a high-quality visual context, ensuring that the context reliably specifies the target identity. The single-identity recognition task then asks whether the named housemate appears in a query image from Day 6 or 7. This temporal gap makes the task a test of persistent identity recognition rather than near-duplicate matching, yielding 250 queries. We further evaluate whether such identity context can support higher-level social reasoning using the RelationQA subset from EgoLifeQA~\cite{egolife}. RelationQA already defines four-way questions over Jake's egocentric video, but does not provide the curated identity context needed to isolate visual context utilization. We therefore pair each question with our manually selected reference galleries for the household members, requiring the model to resolve participant identities from personal visual context before answering the social question. This task contains 125 queries.

\paragraph{Objects.} The objects axis evaluates whether a model can ground a personally specified object instance, rather than merely recognize an object category. We build this axis from Ego4D Visual Queries~\cite{ego4d} by using its object tracks and bounding-box annotations as raw visual evidence, then redesigning them into Personal VCL context-query tasks. Across eight everyday object categories, bottle, bowl, box, bucket, chair, container, cup, and pliers, we form reference galleries for individual object instances. In instance recognition, the model is given one to five reference images of a target object, such as ``my bottle,'' and must decide whether the same instance appears in a query video. Negative queries are drawn from other instances of the same category whenever possible, forcing instance-level discrimination rather than category recognition. This task contains 660 queries over 89 object instances. In spatial detection, we use the same personalized reference setup but ask the model to localize the target instance in a query frame, using Ego4D's bounding-box annotations for evaluation. This yields 138 positive localization queries over 88 object instances. Together, the two tasks turn object-level egocentric annotations into a test of personalized object grounding: recognizing and localizing the specific object tied to the user.

\paragraph{Behavior.} The behavior axis targets a form of personal knowledge that is difficult to capture with static concepts: how a user performs a routine action. We redesign CaptainCook4D~\cite{captaincook4d} into this Personal VCL setting by constructing same-user, same-step context-query pairs. For procedural error detection, the context clips show the participant's correct execution of a step, while the query clip shows the same participant performing that step again. The model must decide whether the query deviates from the demonstrated baseline. Because the context and query involve the same user and the same procedural step, the task focuses on personalized deviation detection rather than generic mistake recognition. This task contains 370 balanced binary queries. For procedural question answering, a human annotator compares the same context and query videos and identifies fine-grained attributes that remain consistent or differ in the user's execution. We formulate these annotations into 98 four-way multiple-choice questions (MCQ), challenging the model to tell how an execution changes relative to personal visual context.

\paragraph{\codex{EgoWearer identification}.} This final task asks whether an LMM can recognize the continuity of the camera wearer from first-person visual experience. EgoLife~\cite{egolife} provides a particularly strong basis for this task because it records six subjects over multiple days in the same shared apartment, producing natural overlap in scenes, activities, and household objects. Using the dataset's action annotations, we design each example around five reference clips of one wearer performing a particular action. The query clip shows either the same wearer or another individual performing the same action, and the model must decide whether the query wearer matches the reference wearer. Because negatives are drawn from another person doing the same action in an overlapping environment, scene and action cues are intentionally controlled; the model must instead rely on fine-grained identity evidence such as hands, clothing, personal objects, and motion patterns. We build 614 queries spanning 15 daily actions, with 96 manually curated behavior-centric queries whose context clips exhibit clean and stable person-specific motion patterns. The resulting task provides a demanding testbed for multi-context aggregation and fine-grained egocentric identity reasoning.

\begin{figure*}[!t]
    \centering
    \includegraphics[width=\textwidth]{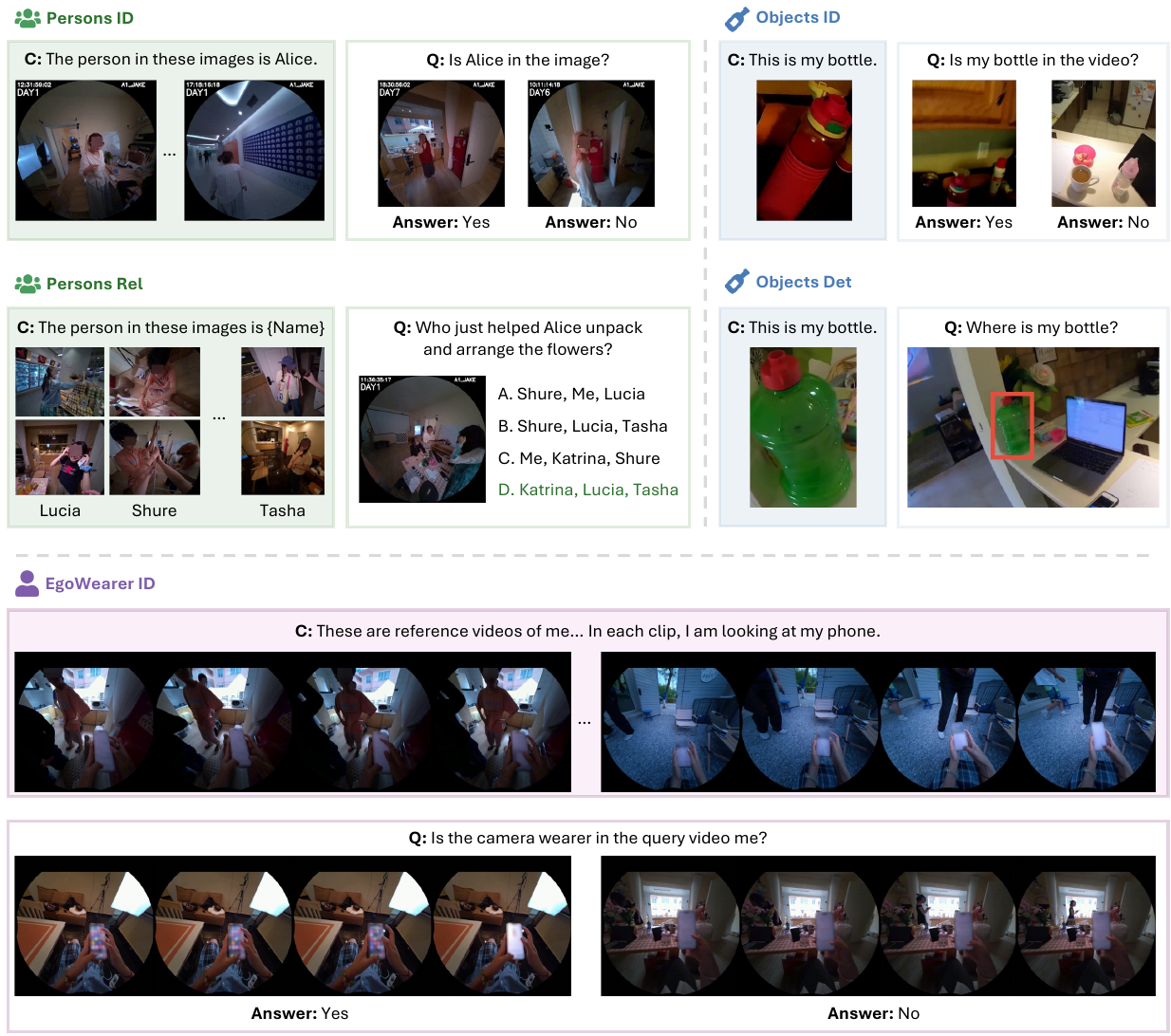}
    \caption{Examples from the Persons, Objects, and EgoWearer identification tasks in Personal-VCL-Bench, built from EgoLife~\cite{egolife} and Ego4D~\cite{ego4d}. C denotes context, and Q denotes the query.
    }
    \label{fig:supp-benchmark-person-object}
\end{figure*}

\begin{figure*}[!t]
    \centering
    \includegraphics[width=\textwidth]{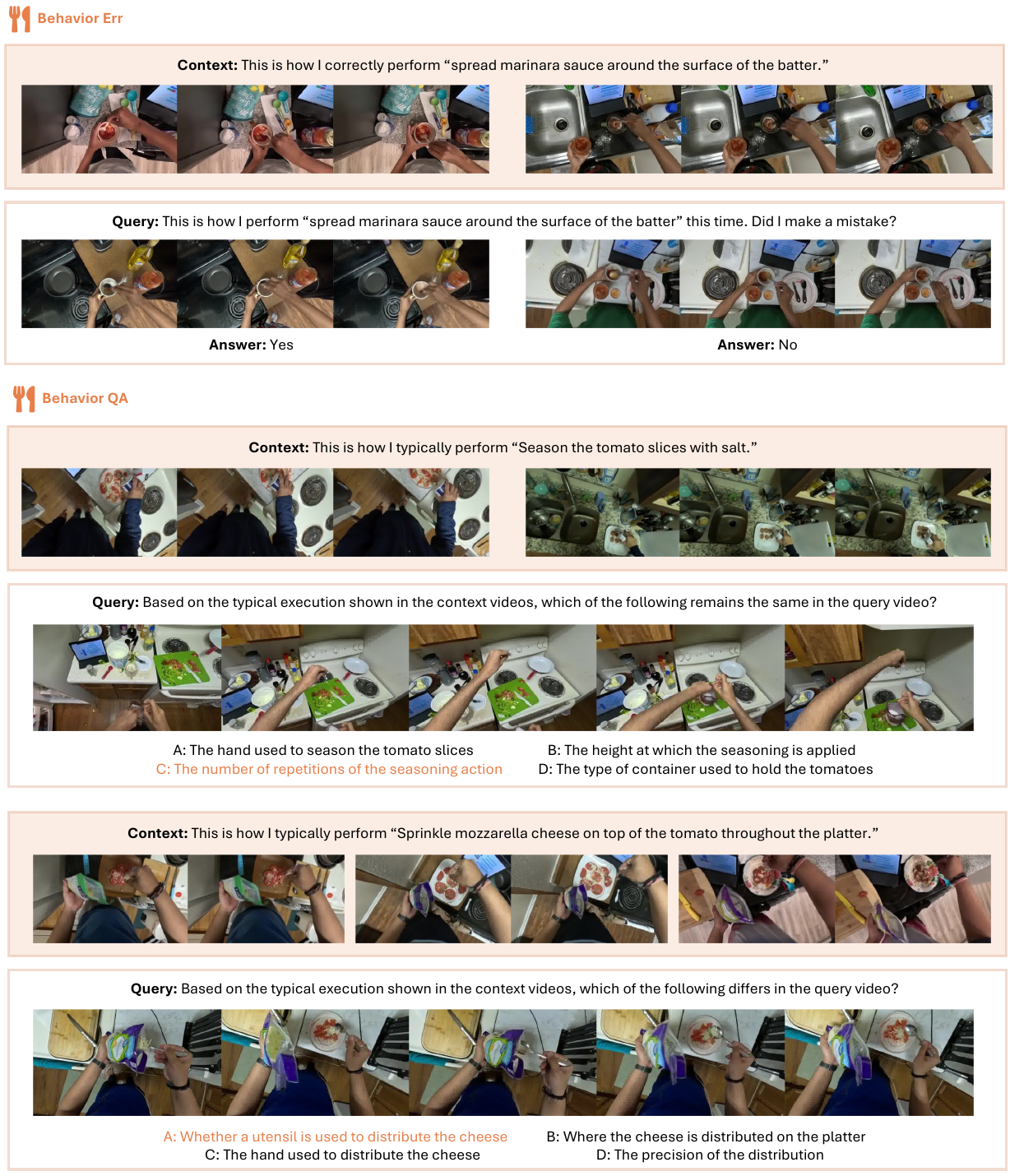}
    \caption{Examples from the Behavior tasks in Personal-VCL-Bench, built from CaptainCook4D~\cite{captaincook4d}.
    }
    \label{fig:cc4d_beherr}
\end{figure*}

\subsection{Evaluation}
We choose metrics according to the output format of each task. For binary recognition tasks, PerID, ObjID, BehErr, and EgoID, we report macro accuracy, averaging Yes and No accuracy so that class imbalance does not reward majority-class predictions. For four-way reasoning tasks, PerRel and BehQA, we report standard multiple-choice accuracy, counting invalid or unparseable outputs as incorrect. For personalized object localization, ObjDet, we evaluate the predicted bounding box by intersection-over-union and report Acc@IoU $\geq$ 0.5. For EgoID, we report a single score over the full 614-query evaluation set: class accuracies are computed after pooling the general and behavior-centric subsets, then macro-averaged.

\section{Extended Diagnostic Results}
\label{sec:supp-diagnostic}

\begin{table}[!th]
\centering
\bencheval
\providecommand{\visup}[2]{#1\rlap{\,\textsuperscript{\textcolor{green!45!black}{$\boldsymbol{\uparrow}$}}}}
\providecommand{\visdown}[2]{#1\rlap{\,\textsuperscript{\textcolor{red!65!black}{$\boldsymbol{\downarrow}$}}}}
\providecommand{\vistie}[2]{#1\rlap{\,\textsuperscript{\textcolor{black!55}{$\boldsymbol{\sim}$}}}}
\caption{Full diagnostic results on Personal-VCL-Bench across five context regimes for all seven evaluated LMMs (per-task accuracy, \%); extends Tab.~\ref{tab:investigation}. \textbf{Bold} = best within a model's five rows; \underline{underline} = best across all models. A random baseline is not directly applicable to the continuous bounding box localization required for ObjDet (Acc@IoU$\geq$0.5) and is marked as `--'. In visual-context rows, cell shading indicates whether visual context improves, hurts, or ties against the matched language-context row. In Visual-ctx ($k_{\max}$), superscript markers indicate whether increasing visual context from $k{=}1$ to $k_{\max}$ improves, drops, or ties.}
\label{tab:investigation-full}
\begin{tabular}{l*{6}{N}E}
\benchheader
Random guess & 50.00 & 25.00 & 50.00 & -- & 50.00 & 25.00 & 50.00 \\
\multicolumn{8}{c}{Qwen3-VL-8B~\cite{qwen3vl}} \\
No-context              & 50.50 & 36.00 & 49.06 & 68.12 & 49.46 & 19.39 & 49.83 \\
Language-ctx ($k{=}1$)    & 51.75 & 29.60 & 65.59 & 76.09 & \textbf{57.57} & \textbf{24.49} & 52.57 \\
Language-ctx ($k_{\max}$) & 62.75 & \textbf{48.80} & 67.25 & \textbf{77.54} & 54.86 & 22.45 & \textbf{54.53} \\
Visual-ctx ($k{=}1$)      & \cellcolor{cellgreen}62.75 & \cellcolor{cellgreen}46.40 & \cellcolor{cellgreen}70.99 & \cellcolor{cellgreen}\textbf{77.54} & \cellcolor{cellgray}\textbf{57.57} & \cellcolor{cellred}21.43 & \cellcolor{cellred}50.00 \\
Visual-ctx ($k_{\max}$)   & \cellcolor{cellgreen}\visup{\textbf{85.25}}{} & \cellcolor{cellred}\vistie{46.40}{} & \cellcolor{cellgreen}\visup{\textbf{73.67}}{} & \cellcolor{cellgray}\vistie{76.81}{} & \cellcolor{cellgray}\visdown{55.14}{} & \cellcolor{cellgray}\visup{22.45}{} & \cellcolor{cellred}\vistie{50.00}{} \\
\benchmidrule
\multicolumn{8}{c}{Qwen3-VL-32B~\cite{qwen3vl}} \\
No-context              & 49.25 & 32.80 & 48.47 & 61.59 & 48.38 & \textbf{34.69} & 49.65 \\
Language-ctx ($k{=}1$)    & 50.00 & 34.40 & 70.29 & \textbf{78.26} & \textbf{51.08} & 29.59 & 52.49 \\
Language-ctx ($k_{\max}$) & 70.25 & 49.60 & 71.02 & 76.09 & 49.73 & 26.53 & 55.34 \\
Visual-ctx ($k{=}1$)      & \cellcolor{cellgreen}67.50 & \cellcolor{cellgreen}43.20 & \cellcolor{cellgreen}78.67 & \cellcolor{cellred}39.86 & \cellcolor{cellgray}50.54 & \cellcolor{cellred}26.53 & \cellcolor{cellgray}53.02 \\
Visual-ctx ($k_{\max}$)   & \cellcolor{cellgreen}\visup{\textbf{85.75}}{} & \cellcolor{cellgreen}\visup{\textbf{56.00}}{} & \cellcolor{cellgreen}\visup{\textbf{79.99}}{} & \cellcolor{cellred}\visdown{33.33}{} & \cellcolor{cellgray}\visdown{48.92}{} & \cellcolor{cellred}\visdown{24.49}{} & \cellcolor{cellgreen}\visup{\underline{\textbf{62.52}}}{} \\
\benchmidrule
\multicolumn{8}{c}{InternVL-3.5-8B~\cite{internvl35}} \\
No-context              & 54.25 & 33.60 & 49.34 & \textbf{19.57} & 51.08 & 11.22 & 47.83 \\
Language-ctx ($k{=}1$)    & 54.50 & 36.00 & 62.86 & 13.77 & 53.24 & \textbf{19.39} & 53.36 \\
Language-ctx ($k_{\max}$) & 59.50 & 36.00 & \textbf{63.05} & 14.49 & 52.97 & 17.35 & \textbf{53.92} \\
Visual-ctx ($k{=}1$)      & \cellcolor{cellgray}54.75 & \cellcolor{cellgreen}\textbf{39.20} & \cellcolor{cellred}50.29 & \cellcolor{cellred}10.87 & \cellcolor{cellgreen}\textbf{54.86} & \cellcolor{cellred}12.24 & \cellcolor{cellred}50.25 \\
Visual-ctx ($k_{\max}$)   & \cellcolor{cellgreen}\visup{\textbf{62.50}}{} & \cellcolor{cellgreen}\vistie{38.40}{} & \cellcolor{cellred}\vistie{50.29}{} & \cellcolor{cellred}\visdown{7.25}{} & \cellcolor{cellgreen}\vistie{54.05}{} & \cellcolor{cellred}\visdown{8.16}{} & \cellcolor{cellred}\vistie{49.83}{} \\
\benchmidrule
\multicolumn{8}{c}{InternVL-3.5-38B~\cite{internvl35}} \\
No-context              & 53.75 & 38.40 & 49.95 & \textbf{13.77} & 50.00 & 22.45 & 50.00 \\
Language-ctx ($k{=}1$)    & 54.50 & 37.60 & 65.41 & 10.87 & \textbf{52.97} & 28.57 & 52.80 \\
Language-ctx ($k_{\max}$) & 75.75 & \textbf{50.40} & \textbf{65.76} & 10.14 & 51.35 & \textbf{31.63} & \textbf{53.78} \\
Visual-ctx ($k{=}1$)      & \cellcolor{cellgreen}77.25 & \cellcolor{cellgreen}42.40 & \cellcolor{cellred}52.02 & \cellcolor{cellgreen}12.32 & \cellcolor{cellred}50.54 & \cellcolor{cellred}17.35 & \cellcolor{cellred}49.84 \\
Visual-ctx ($k_{\max}$)   & \cellcolor{cellgreen}\visup{\textbf{80.50}}{} & \cellcolor{cellred}\visup{47.20}{} & \cellcolor{cellred}\vistie{51.79}{} & \cellcolor{cellgray}\visdown{9.42}{} & \cellcolor{cellred}\vistie{50.00}{} & \cellcolor{cellred}\visup{19.39}{} & \cellcolor{cellred}\vistie{50.17}{} \\
\benchmidrule
\multicolumn{8}{c}{Gemma-4-31B~\cite{gemma4}} \\
No-context              & 49.25 & 22.40 & 51.08 & 57.97 & 52.43 & 29.59 & 50.13 \\
Language-ctx ($k{=}1$)    & 57.50 & 20.80 & 78.33 & \textbf{72.46} & 54.86 & 33.67 & 54.52 \\
Language-ctx ($k_{\max}$) & \textbf{78.00} & \textbf{49.60} & \textbf{78.75} & 71.74 & 58.11 & \textbf{34.69} & 52.99 \\
Visual-ctx ($k{=}1$)      & \cellcolor{cellgreen}65.50 & \cellcolor{cellgreen}30.40 & \cellcolor{cellred}73.26 & \cellcolor{cellred}63.77 & \cellcolor{cellgreen}\textbf{59.46} & \cellcolor{cellred}32.65 & \cellcolor{cellgreen}\textbf{56.66} \\
Visual-ctx ($k_{\max}$)   & \cellcolor{cellred}\visup{75.75}{} & \cellcolor{cellred}\visup{46.40}{} & \cellcolor{cellred}\visup{75.34}{} & \cellcolor{cellred}\visup{65.22}{} & \cellcolor{cellgreen}\vistie{\textbf{59.46}}{} & \cellcolor{cellred}\visdown{27.55}{} & \cellcolor{cellgreen}\visdown{55.36}{} \\
\benchmidrule
\multicolumn{8}{c}{Gemini-3-Flash~\cite{gemini3flash}} \\
No-context              & 55.75 & 42.40 & 49.32 & 64.49 & 57.84 & 37.76 & 49.47 \\
Language-ctx ($k{=}1$)    & 69.25 & 37.60 & 80.55 & 79.71 & 55.41 & 45.92 & 53.99 \\
Language-ctx ($k_{\max}$) & 81.75 & \textbf{59.20} & 82.35 & 78.99 & 54.59 & 48.98 & 55.26 \\
Visual-ctx ($k{=}1$)      & \cellcolor{cellred}67.50 & \cellcolor{cellgreen}44.00 & \cellcolor{cellgreen}82.21 & \cellcolor{cellgreen}\underline{\textbf{85.51}} & \cellcolor{cellgreen}\underline{\textbf{68.92}} & \cellcolor{cellgreen}47.96 & \cellcolor{cellred}51.99 \\
Visual-ctx ($k_{\max}$)   & \cellcolor{cellgreen}\visup{\textbf{88.00}}{} & \cellcolor{cellred}\visup{54.40}{} & \cellcolor{cellgray}\vistie{\textbf{82.47}}{} & \cellcolor{cellgreen}\visdown{82.61}{} & \cellcolor{cellgreen}\visdown{67.03}{} & \cellcolor{cellgreen}\visup{\textbf{61.22}}{} & \cellcolor{cellgray}\visup{\textbf{55.74}}{} \\
\benchmidrule
\multicolumn{8}{c}{Gemini-3.1-Pro~\cite{gemini31}} \\
No-context              & 54.50 & 43.20 & 51.64 & 72.46 & 66.22 & 35.71 & 49.35 \\
Language-ctx ($k{=}1$)    & 52.75 & 44.00 & 67.54 & 82.61 & 58.38 & 47.96 & 48.73 \\
Language-ctx ($k_{\max}$) & 73.50 & 45.60 & 78.28 & \textbf{83.33} & 56.22 & 51.02 & 49.75 \\
Visual-ctx ($k{=}1$)      & \cellcolor{cellgreen}83.75 & \cellcolor{cellgreen}50.40 & \cellcolor{cellgreen}\underline{\textbf{83.63}} & \cellcolor{cellred}55.07 & \cellcolor{cellgreen}64.32 & \cellcolor{cellgreen}61.22 & \cellcolor{cellgreen}55.90 \\
Visual-ctx ($k_{\max}$)   & \cellcolor{cellgreen}\visup{\underline{\textbf{88.75}}}{} & \cellcolor{cellgreen}\visup{\underline{\textbf{60.00}}}{} & \cellcolor{cellgreen}\visdown{82.14}{} & \cellcolor{cellred}\visdown{50.00}{} & \cellcolor{cellgreen}\visup{\textbf{68.11}}{} & \cellcolor{cellgreen}\visup{\underline{\textbf{65.31}}}{} & \cellcolor{cellgreen}\visup{\textbf{58.16}}{} \\
\benchbottomrule
\end{tabular}
\end{table}

We provide the full diagnostic table in Tab.~\ref{tab:investigation-full}, extending the representative results in the main paper to all seven LMMs. The evaluation follows the same five-regime protocol for every model: no-context, language-context with one item, language-context with all items, visual-context with one item, and visual-context with all items.

The results support three conclusions. First, Personal-VCL-Bench largely resists generic pretrained knowledge. Identity-style tasks such as PerID, ObjID, and EgoID remain close to chance for most models, while behavior and relation tasks can contain partial signal in the query itself. This makes the subsequent context comparisons more important: they measure whether language or visual context helps models move beyond generic query understanding toward genuine personal visual context utilization. Second, the modality comparison exposes a gap in native visual context utilization. Visual context can be powerful, particularly for person and object identity, but it is not reliably dominant; language descriptions often match or outperform raw visual evidence despite discarding visual detail. Third, the $k{=}1$ versus $k_{\max}$ comparison shows that current LMMs do not reliably aggregate personal context. More reference images or clips can help when the task resembles visual matching, but the benefit becomes unstable for behavior, localization, and \codex{EgoWearer identification}. This pattern is central to our motivation: Personal VCL requires not only retrieving relevant visual history, but also organizing and selecting from it in a way that current prompt concatenation does not provide.

\clearpage

\section{Implementation Details}
\label{sec:supp-implementation}

\subsection{Context prompting baselines}
We evaluate three direct prompting baselines that vary only in how personal context is represented. The no-context baseline receives only the query media and the task question. The visual-context baseline presents the selected reference images or clips before the query. The language-context baseline first asks the same model to describe each reference item, then presents the task using those descriptions in place of the raw reference media. For both context baselines, we report a single-reference condition and a full-reference condition, denoted $k{=}1$ and $k_{\max}$. Across these settings, the task question is unchanged; the comparison isolates whether the model can use personal context, and whether that context is more effective as raw visual evidence or as text.
We uniformly sample 16 frames for all video inputs, except for Gemini models, which process native video via their API. These identical settings are applied when generating language-context descriptions.

\paragraph{Visual context baseline prompts.}
The task-specific visual-context prompts are shown below.
\begin{promptquote}
\promptaxislabel{axispersons}{\faUsers~Persons}\\
\smallskip
For single-identity recognition, the context prompt is ``The person in these images is \texttt{[person]}'' followed by the query ``Is \texttt{[person]} in this image?'' For relationship reasoning, each household member is introduced with the same identity prompt, and the query is the RelationQA question.

\medskip
\promptaxislabel{axisobjects}{\faWineBottle~Objects}\\
\smallskip
For instance recognition and spatial detection, the shared context prompt is ``This is my \texttt{[object]}.'' The recognition query asks ``Is my \texttt{[object]} in the video?'' The detection query asks ``Where is my \texttt{[object]}? Return the bounding box coordinates.''

\medskip
\promptaxislabel{axisbehavior}{\faUtensils~Behavior}\\
\smallskip
For procedural error detection, the context prompt is ``This is how I correctly perform \texttt{[step]}.'' The query asks ``This is how I perform \texttt{[step]} this time. Did I make a mistake?'' For procedural question answering, the context prompt is ``This is how I typically perform \texttt{[step]}.'' The query is the human-written multiple-choice question about which aspect differs from, or remains consistent with, the context execution, followed by four answer choices.

\medskip
\promptaxislabel{axisego}{\faUser~EgoWearer}\\
\smallskip
For \codex{EgoWearer identification}, the context prompt is ``These are reference videos of me, recorded with a head-mounted fisheye camera in a shared apartment with my housemates. In each clip I am \texttt{[action]}.'' The query asks: ``All reference and query clips share the same camera rig, the same apartment, the same housemates, and the same action, so none of those are evidence of identity. Look for subtler, person-specific details that remain. Is the camera wearer in the query video me?''
\end{promptquote}

\paragraph{Language context baseline prompts.}
The description prompts are shown below.
\begin{promptquote}
\promptaxislabel{axispersons}{\faUsers~Persons}\\
\smallskip
For persons, the description prompt is ``These are images of \texttt{[person]}. Describe \texttt{[person]}'s physical appearance in one paragraph, focusing only on features that would help identify them: hair color and style, clothing, accessories such as glasses, jewelry, hats, or bags, body type, and any other distinguishing visual features. Do not describe the background, setting, or activities. Starting with `\texttt{[person]} is'.'' The generated descriptions replace the reference images as the identity context for both single-identity recognition and relationship reasoning.

\medskip
\promptaxislabel{axisobjects}{\faWineBottle~Objects}\\
\smallskip
For objects, the description prompt is ``This is my \texttt{[object]}. Describe my \texttt{[object]} in one paragraph. Starting with `It is'.'' The generated descriptions replace the reference images as the object context, framed as ``This is my \texttt{[object]}. \texttt{[description]}.''

\medskip
\promptaxislabel{axisbehavior}{\faUtensils~Behavior}\\
\smallskip
For behavior, the description prompt is ``Describe how I perform \texttt{[step]}.'' The generated descriptions replace the reference clips as the procedural context, framed as ``This is how I correctly perform \texttt{[step]}: \texttt{[description]}.''

\medskip
\promptaxislabel{axisego}{\faUser~EgoWearer}\\
\smallskip
For \codex{EgoWearer identification}, each reference clip is described with ``This is a first-person video clip of me \texttt{[action]}. Describe my appearance and my movements in detail.'' The generated descriptions replace the reference clips as the wearer context, using the same apartment-and-action lead-in as the visual-context baseline, followed by ``Below is a description of one such clip:'' or ``Below are descriptions of \texttt{[n]} such clips:'' and the per-clip descriptions.
\end{promptquote}

\subsection{Agentic Context Bank}
\paragraph{Stage I: bank construction.}
In the multi-clip \codex{EgoWearer identification} setting, each context clip is sampled into 16 uniformly spaced frames before the LMM extracts reusable personal cues about the camera wearer. Static cues, including appearance and owned objects, are grounded to a single supporting frame; motion cues are grounded to a temporal span.
The extraction prompt is shown below.

\begin{promptquote}
\promptsteplabel{Extraction.}\\
\smallskip
``You are observing an egocentric (first-person) video clip from a head-mounted camera. The video has been sampled into \texttt{[N]} frames, labeled in temporal order. Extract distinctive cues about the camera wearer---observations that capture who this individual is and could be reused to reason about them in other contexts.''

\smallskip
\promptitemlabel{Static cues:} what the wearer's body, clothing, wearables, or owned objects look like. Anchor each cue to one best frame.

\smallskip
\promptitemlabel{Motion cues:} how the wearer moves and acts. Anchor each cue to one temporal span.

\smallskip
\promptitemlabel{Rules:} one entry per distinct cue; do not bundle unrelated attributes; emit only cues grounded in a specific frame or span; prefer specific over vague descriptions; avoid restating the action label.
\end{promptquote}

After extraction, candidates are reconciled with the current bank separately for each memory type. Each stored entry maintains a stable identifier, a descriptor, visual evidence, and support counts. Revisions are accepted only after a separate visual verification prompt.
The merge and revision-verification prompts are shown below.

\begin{promptquote}
\promptsteplabel{Merge.}\\
\smallskip
``You are reconciling new \texttt{[category]} cues against an existing memory of a person.''

\smallskip
\promptitemlabel{Existing entries:} \texttt{[active bank entries in this category]}.

\smallskip
\promptitemlabel{New candidates:} \texttt{[c\_001]}, \texttt{[c\_002]}, $\ldots$, each with its supporting frame or temporal span.

\smallskip
\promptitemlabel{Output:} for each useful candidate, choose \texttt{ADD}, \texttt{CONFIRM}, \texttt{REVISE}, or \texttt{RETRACT}; silently drop redundant noise. Use \texttt{REVISE} only when the candidate refines the same underlying attribute.\\

\medskip
\promptsteplabel{Revision verification.}\\
\smallskip
``You proposed the following REVISE operations. For each one, I am showing the visual evidence from both the existing entry and the new candidate. Verify that they refer to the same \texttt{[category]} attribute of the same person. If they do, confirm the REVISE; otherwise, withdraw it.''
\end{promptquote}

\paragraph{Stage II: query-adaptive evidence.}
At query time in this same wearer-recognition setting, the active bank is rendered as text grouped by appearance, owned objects, and behavior. The first call receives this text view together with 16 uniformly sampled query frames. If visual grounding is requested, the second call attaches evidence only for the requested entries: a supporting frame for static entries and sampled span frames for behavior entries. Across the reported agentic experiments, each requested behavior span is represented by up to four uniformly spaced frames for Gemini/GPT and by its two endpoints for Gemma.
The two query-time prompts are shown below.

\begin{promptquote}
\promptsteplabel{Call 1: text triage.}\\
\smallskip
``These are reference videos of me, recorded with a head-mounted fisheye camera in a shared apartment with my housemates. In each clip I am \texttt{[action]}. Below is a structured memory of me, built from those clips:

\smallskip
\texttt{[bank text]}

\smallskip
Now here are frames from the query clip.''

\smallskip
\promptitemlabel{Question:} ``All reference and query clips share the same camera rig, the same apartment, the same housemates, and the same action, so none of those are evidence of identity. Look for subtler, person-specific details that remain. Is the camera wearer in the query video me?''

\smallskip
\promptitemlabel{Decision:} If you can decide from the text claims alone, answer Yes or No; otherwise, request the bank entries whose visual evidence would help you decide.\\

\medskip
\promptsteplabel{Call 2: selective visual verification.}\\
\smallskip
``Here is the same structured memory of me (reference videos where I am \texttt{[action]}), now with visual evidence attached for the entries you requested:

\smallskip
\texttt{[bank text, with frames inlined only for requested entries]}

\smallskip
Visual evidence has been attached for the entries you requested: \texttt{[entry ids]}. Compare the attached visual evidence to the query video. Is the camera wearer in the query video me? Answer Yes or No and report the decisive entries.''
\end{promptquote}

\paragraph{Task-wise extensions.}
The preceding prompts describe the full two-stage Agentic Context Bank as instantiated for \codex{EgoWearer identification}. For the task-wise results in Tab.~\ref{tab:gemma-taskwise}, we adapt the bank to the structure of each benchmark axis. For entity-centric tasks in the persons and objects axes (PerID, PerRel, ObjID, ObjDet), there is little temporal structure to merge: each context item is already a reference observation of a known person or object. We therefore use a per-item description bank and focus on stage-II query-adaptive evidence selection. The model first triages the text bank, then selectively requests the corresponding visual evidence when the descriptions alone are insufficient. For the behavior axis, the relevant context is procedural rather than instance-level. We construct a phase-structured behavior bank from the reference executions and focus on evaluating the resulting stage-I bank.
The task-wise prompt templates are shown below.

\paragraph{Compute resources.} All reported experiments use frozen models at inference time. For local LMMs, we run inference on an NVIDIA H200 GPU cluster, with each inference job allocated one H200 GPU node. Proprietary models, including the Gemini and OpenAI families, are accessed via their official APIs.

\begin{promptquote}
\promptaxislabel{axispersons}{\faUsers~Persons} \quad \promptaxislabel{axisobjects}{\faWineBottle~Objects}\\
\smallskip
\promptitemlabel{Call 1:} ``Here are text descriptions of \texttt{[reference images/clips]}:

\smallskip
\texttt{[e\_001]} \texttt{[description 1]}\\
\texttt{[e\_002]} \texttt{[description 2]}\\
$\cdots$

\smallskip
Now here is the query: \texttt{[task question]}. Answer directly from the text descriptions, or request the specific entries whose visual evidence you need.''

\smallskip
\promptitemlabel{Call 2:} ``Here are the reference images or clips you requested. \texttt{[requested visual evidence]} Now, with this visual evidence alongside the query, answer the question: \texttt{[task question]}.''

\medskip
\promptaxislabel{axisbehavior}{\faUtensils~Behavior}\\
\smallskip
\promptitemlabel{BehErr:} ``Here is a record of how I perform \texttt{[step]}, broken into temporal phases (in order):

\smallskip
\texttt{[e\_001]} \texttt{[phase 1]}\\
\texttt{[e\_002]} \texttt{[phase 2]}\\
$\cdots$

\smallskip
Now here are frames from a new clip where I perform the same action. Did I make a mistake? Answer Yes or No.''

\smallskip
\promptitemlabel{BehQA:} ``Here is a record of how I perform \texttt{[step]}, broken into temporal phases (in order):

\smallskip
\texttt{[e\_001]} \texttt{[phase 1]}\\
\texttt{[e\_002]} \texttt{[phase 2]}\\
$\cdots$

\smallskip
Detailed description of how I typically perform this action across the reference clips:
\texttt{[reference description]}

\smallskip
\texttt{[MCQ question]}''
\end{promptquote}

\section{Additional Results and Analysis}
\label{sec:supp-additional}

\paragraph{Stage-I ablation.}
\begin{table}[!htbp]
\centering
\bencheval
\caption{Effect of stage-I bank construction on \codex{EgoWearer Identification}. With Gemma-4-31B~\cite{gemma4} as the query model, we compare flat visual-text memory, the full structured bank, an order-perturbed bank, and a bank constructed by a stronger builder. The full bank improves over flat memory, is stable to reference order, and benefits from stronger construction.}
\label{tab:stage1-ablation}
\begin{tabular}{l>{\centering\arraybackslash}p{2.6cm}>{\centering\arraybackslash}p{2.4cm}}
\toprule
Memory construction & Builder & \mbox{\textcolor{axisego}{\faUser~EgoWearer}} \\
\midrule
Flat visual-text memory & Gemma-4-31B & 51.11 \\
Full bank & Gemma-4-31B & 61.60 \\
Full bank (\codex{permuted} order) & Gemma-4-31B & 62.11 \\
Full bank & Gemini-3-Flash & 65.53 \\
\bottomrule
\end{tabular}
\end{table}

Tab.~\ref{tab:stage1-ablation} separates the effect of stage-I memory construction from the stage-II evidence-selection mechanism analyzed in the main paper. We consider three variants. First, we replace the structured bank with a flat visual-text memory, where each reference clip is represented by an independent description without extract-and-merge. Second, we perturb the order in which reference clips are processed during bank construction, testing whether the sequential update process is sensitive to context order. Third, we substitute the Gemma-built bank with a Gemini-built bank while still using Gemma for the final query-time decision. When the memory is reduced to flat per-clip descriptions, the same query-adaptive procedure obtains 51.11\%, indicating that selective access alone is not sufficient if the underlying memory remains unstructured. The perturbed-order bank performs similarly at 62.11\%, suggesting that the gain is not an ordering artifact. The Gemini-built bank further improves to 65.53\%, indicating that stronger stage-I memory construction can directly benefit downstream Personal VCL.

\paragraph{Bank analysis.}
\begin{table}[!htbp]
\centering
\bencheval
\caption{Stage-I bank-construction statistics on \codex{EgoWearer Identification}. Cues/clip counts extracted candidate cues. Entries reports the mean final bank size, with A/O/B denoting appearance, owned-object, and behavior entries. Compression is the ratio between final bank entries and extracted candidates. Revision ops are the share of merge decisions that update an existing entry, and updated entries are final entries confirmed or revised after creation.}
\label{tab:bank-construction}
\begin{tabular}{lcccccc}
\toprule
Builder & Cues/clip & Entries & A/O/B & Comp. & Rev. ops & Updated \\
\midrule
Gemini-3-Flash & 4.44 & 13.8 & 6.0/3.1/4.7 & 0.62 & 35.9\% & 24.1\% \\
Gemma-4-31B & 2.95 & 8.2 & 3.8/1.5/3.0 & 0.56 & 25.2\% & 18.3\% \\
GPT-5.4-mini & 6.92 & 18.4 & 6.7/5.7/5.9 & 0.53 & 38.1\% & 26.5\% \\
\bottomrule
\end{tabular}
\end{table}

\begin{table}[!htbp]
\centering
\bencheval
\caption{Stage-II query-time evidence statistics on \codex{EgoWearer Identification}. Visual requests are the fraction of queries for which the model asks to inspect stored visual evidence. Requested and decisive entries are averaged over visual-evidence queries. A/O/B reports the percentage distribution over appearance, owned-object, and behavior entries.}
\label{tab:bank-query}
\begin{tabular}{lccccc}
\toprule
Builder & Vis. req. & Req. & Req. A/O/B & Dec. & Dec. A/O/B \\
\midrule
Gemini-3-Flash & 95.8\% & 3.63 & 81.8/15.1/3.0 & 2.67 & 79.2/15.2/5.5 \\
Gemma-4-31B & 96.1\% & 2.86 & 85.9/11.2/2.9 & 1.78 & 80.7/13.3/6.0 \\
GPT-5.4-mini & 2.1\% & 8.77 & 60.5/16.7/22.8 & 4.31 & 85.7/5.4/8.9 \\
\bottomrule
\end{tabular}
\end{table}

Tab.~\ref{tab:bank-construction} summarizes the stage-I banks constructed for \codex{EgoWearer identification}, where each bank is built from five reference clips. Across builders, extraction produces 2.95--6.92 candidate cues per clip, and the merge process compresses them into 8.2--18.4 final entries per bank, corresponding to compression ratios of 0.53--0.62. The final banks remain organized across all three evidence categories: appearance, owned objects, and behavior. The bank construction is not a simple append-only process. Revision operations account for 25--38\% of merge decisions, and 18--27\% of final entries are updated through later confirmation or revision, indicating that the bank accumulates and refines reusable personal evidence across clips.

Tab.~\ref{tab:bank-query} analyzes how the constructed bank is accessed during stage-II querying. For Gemini-3-Flash~\cite{gemini3flash} and Gemma-4-31B~\cite{gemma4}, the model requests stored visual evidence for nearly all queries, while still selecting a small subset of the bank: 3.63 and 2.86 requested entries on average, respectively. The model-reported decisive set is smaller, averaging 2.67 and 1.78 entries. The requested and decisive distributions span appearance, owned-object, and behavioral evidence, showing that the bank supports decisions through multiple forms of personal evidence rather than a single cue type. This distribution closely mirrors human reasoning. We manually label a 90-query subset as a reference point, where human accuracy reaches 84.1\% (excluding ambiguous cases). The human evaluators cited personal objects (46.3\%), appearance (41.5\%), and behavior (12.2\%) as their primary evidence, validating our structural choice to organize the bank around multiple complementary cues.
GPT-5.4-mini~\cite{gpt54} follows a different access pattern, usually deciding from the text view and requesting visual evidence in 2.1\% of queries; when it does request evidence, it inspects a larger set. This strong preference for text-based resolution likely explains why its overall performance gain is smaller compared to the other two backbones in Tab.~\ref{tab:main}. Together, these statistics show that our stage I effectively compresses and refines multi-category personal evidence, and stage II selects a targeted evidence subset for each query.

\paragraph{Failure cases.}
\begin{figure}[!t]
    \centering
    \includegraphics[width=1.0\linewidth]{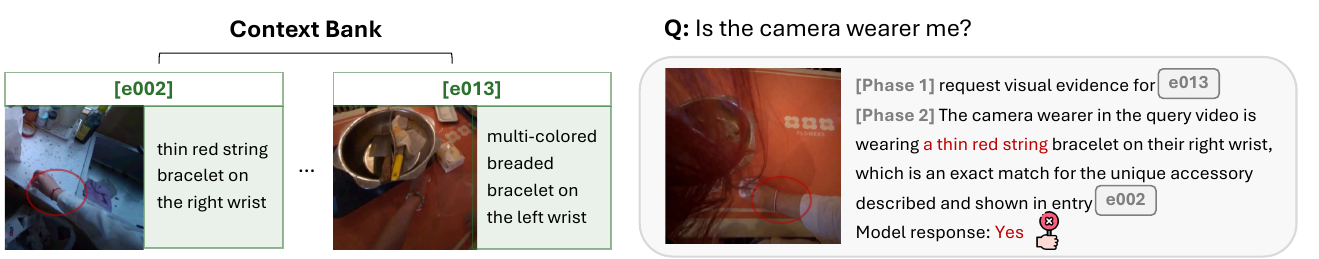}
    \caption{
    Failure case of our Agentic Context Bank. The model requests visual evidence for entry \texttt{e013}, but does not request the raw visual evidence for entry \texttt{e002}, whose text descriptor later drives the final answer. The query wearer has a different bracelet from the one shown in \texttt{e002}; without revisiting the corresponding visual evidence, the model incorrectly treats the descriptor as an exact match and answers Yes.
    }
    \label{fig:failure-case}
\end{figure}

Fig.~\ref{fig:failure-case} illustrates a failure mode of the query-adaptive stage. Although the relevant information exists in the constructed bank, the model fails to request the specific visual evidence required to verify a fine-grained detail. Instead, after retrieving a different bracelet-related entry, it relies solely on the text description of \texttt{e002} and incorrectly infers a match. Because the query frame reveals that the bracelet is visually distinct, the correct answer is No. This example highlights that personalization errors stem 
from inadequate evidence selection at query time.

\section{Broader Impacts}
\label{sec:supp-impact}
This work studies a capability needed for future personalized multimodal assistants: using visual context from a user's own history to answer user-specific questions. If developed responsibly, Personal VCL could make assistants more helpful in everyday settings. It could support remembering personal objects, recognizing familiar people with user consent, comparing a user's current action to their usual routine, or providing individualized feedback in domains such as cooking, rehabilitation, sports, and craft practice. These applications could be especially valuable when generic visual understanding is insufficient because the relevant information is personal rather than part of public world knowledge.

The same capability also creates risks. Personal visual histories are inherently sensitive. They may encode the identity and behavior of the wearer, the presence and actions of bystanders, private spaces, social relationships, and repeated routines. Models that can organize and reason over this information could be used for surveillance, unwanted identification, behavioral profiling, or continuous authentication without meaningful consent. Errors in such systems could also cause harm if they lead to incorrect personalized guidance or misidentify people, objects, or deviations from routine.

Our work is intended as an evaluation benchmark and inference-time study, not as a deployed personal assistant. We use existing research datasets and focus on measuring whether LMMs can use curated personal visual context. Future real-world Personal VCL systems should require informed consent, clear data ownership, strong access control, secure and privacy-preserving storage, and limits on secondary use of personal visual data. Deployment should also consider bystander privacy and provide mechanisms for deletion, auditing, and user control over what visual memory is retained.

\end{document}